\documentclass[lettersize,journal]{IEEEtran}
\usepackage{amsmath,amsfonts}
\usepackage{algorithmic}
\usepackage{algorithm}
\newcommand{\FUNCTION}[2]{%
  \STATE \textbf{def} \textsc{#1} (#2):
}
\newcommand{\ENDFUNCTION}

\usepackage{xcolor}
\usepackage{array}
\usepackage[caption=false,font=normalsize,labelfont=sf,textfont=sf]{subfig}
\usepackage{textcomp}
\usepackage{stfloats}
\usepackage{url}
\usepackage{verbatim}
\usepackage{graphicx}
\usepackage{cite}
\usepackage{booktabs}
\usepackage{multirow}
\usepackage{hyperref}
\IEEEpubid{\begin{minipage}{\textwidth}\ \centering
		Copyright \copyright 2025 IEEE. Personal use of this material is permitted. However, permission to use this material for any other purposes must be obtained from the IEEE by sending an email to pubs-permissions@ieee.org.
\end{minipage}}

\begin{document}

\title{LoDisc: Learning Global-Local Discriminative Features for Self-Supervised Fine-Grained Visual Recognition}

\author{Jialu Shi, Zhiqiang Wei, Jie Nie, and Lei Huang*,~\IEEEmembership{Member,~IEEE}
        % <-this % stops a space
\thanks{This work is supported by the National Natural Science Foundation of China (No. 62472390), Shandong Provincial Natural Science Foundation (ZR2023MF033), National Key R\&D Program of China (2019YFC1408405).}% <-this % stops a space
\thanks{Jialu Shi, Zhiqiang Wei, Jie Nie and Lei Huang are with the Faculty of Information Science and Engineering, Ocean University of China, Qingdao 266100, China (e-mail: shijialu@stu.ouc.edu.cn; weizhiqiang@ouc.edu.cn; niejie@ouc.edu.cn; huangl@ouc.edu.cn).}
\thanks{*Lei Huang is the corresponding author.}
}



\maketitle

\begin{abstract}
The self-supervised contrastive learning strategy has attracted considerable attention due to its exceptional ability in representation learning. However, current contrastive learning tends to learn global coarse-grained representations of the image that benefit generic object recognition, whereas such coarse-grained features are insufficient for fine-grained visual recognition. In this paper, we incorporate subtle local fine-grained feature learning into global self-supervised contrastive learning through a pure self-supervised global-local fine-grained contrastive learning framework. Specifically, a novel pretext task called local discrimination (LoDisc) is proposed to explicitly supervise the self-supervised model's focus toward local pivotal regions, which are captured by a simple but effective location-wise mask sampling strategy. We show that the LoDisc pretext task can effectively enhance fine-grained clues in important local regions and that the global-local framework further refines the fine-grained feature representations of images. Extensive experimental results on different fine-grained object recognition tasks demonstrate that the proposed method can lead to a decent improvement in different evaluation settings. The proposed method is also effective for general object recognition tasks.
\end{abstract}

\begin{IEEEkeywords}
Self-supervised contrastive learning, fine-grained visual recognition, pretext task.
\end{IEEEkeywords}

\section{Introduction}
\IEEEPARstart{I}{n} computer vision, self-supervised learning (SSL) is proposed to learn general representations from unlabeled data and has provided competitive performance of representations against supervised learning in many downstream tasks \cite{1}. Contrastive learning based on contrastive loss \cite{2} functions is the dominant strategy of several recent works on self-supervised learning, which has demonstrated state-of-the-art representation learning capability on generic datasets such as ImageNet \cite{3}. However, compared with generic object recognition (GOR) tasks, fine-grained visual recognition (FGVR) tasks aim to distinguish similar subclasses of a given object category \cite{Intro-tcsvt-1}, such as different models of cars; this requires the neural network to focus more on the subtle differences in fine-grained categories \cite{9, 10, Intro-tcsvt-2}, and discriminative features are localized more predominantly on object parts, e.g., the headlights of a car \cite{11, Intro-tcsvt-3}.

\begin{figure}[t]
  \centering
  \includegraphics[width=0.98\linewidth]{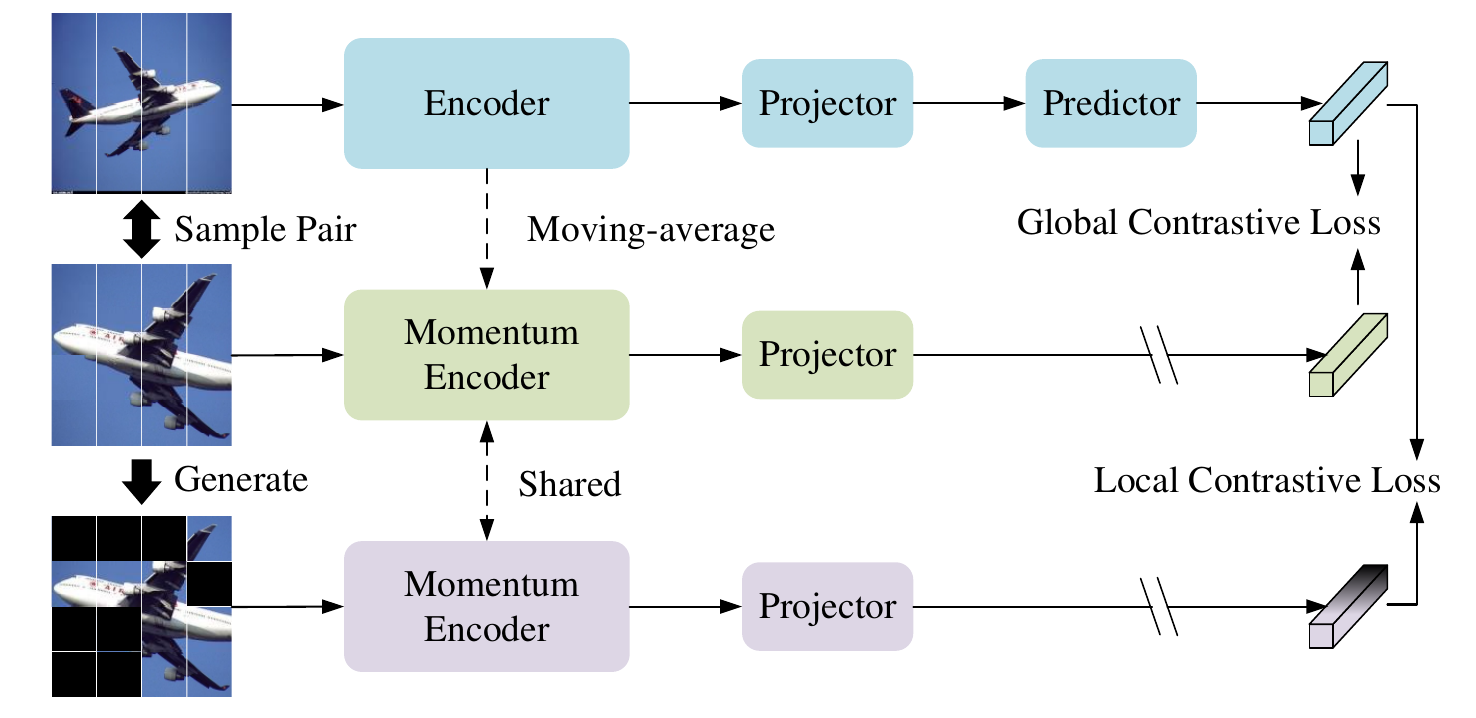}
  \caption{Global-local fine-grained visual recognition multi-branch framework. Sample pairs are augmentations of images, and local contrastive branch's inputs are generated through the augmentation of these images. Global contrastive branches learn global discriminative features based on InstDisc in a self-supervised vision transformer (MoCo v3), which has an encoder and a momentum encoder. Local contrastive branches learn local discriminative features via LoDisc, which shares a momentum encoder.}
  \label{fig:1}
\end{figure}

Recent studies \cite{8} have shown that current contrastive learning tends to learn global coarse-grained features suited for GOR, which are insufficient for FGVR. Specifically, current contrastive learning methods, which are based on instance discrimination (InstDisc) \cite{14}, aim to maximize similar representations between different augmentations of the same image while minimizing the similarity of representations across different images \cite{13}; this enables models to learn global discriminative representations of overall visual similarity but may overlook many delicate details that are beneficial for FGVR \cite{16}. Since differences in fine-grained images are often confined to small and localized regions, learning from a global perspective during unannotated feedback may introduce irrelevant or even distracting features. Additionally, obtaining annotated labels for fine-grained images is challenging due to the need for detailed labeling and the expertise required to accurately distinguish the subtle differences between subclasses. Therefore, focusing on important local regions within a broad global scope, even without annotated information, and further exploring fine-grained features in these regions can effectively aid in accurately capturing fine-grained details; this motivates us to explore methods that encourage self-supervised contrastive learning to further emphasize local parts and uncover subtle distinctions within these local regions.

\IEEEpubidadjcol 

\begin{figure*}
    \subfloat[InstDisc]{
        \includegraphics[width=0.48\linewidth]{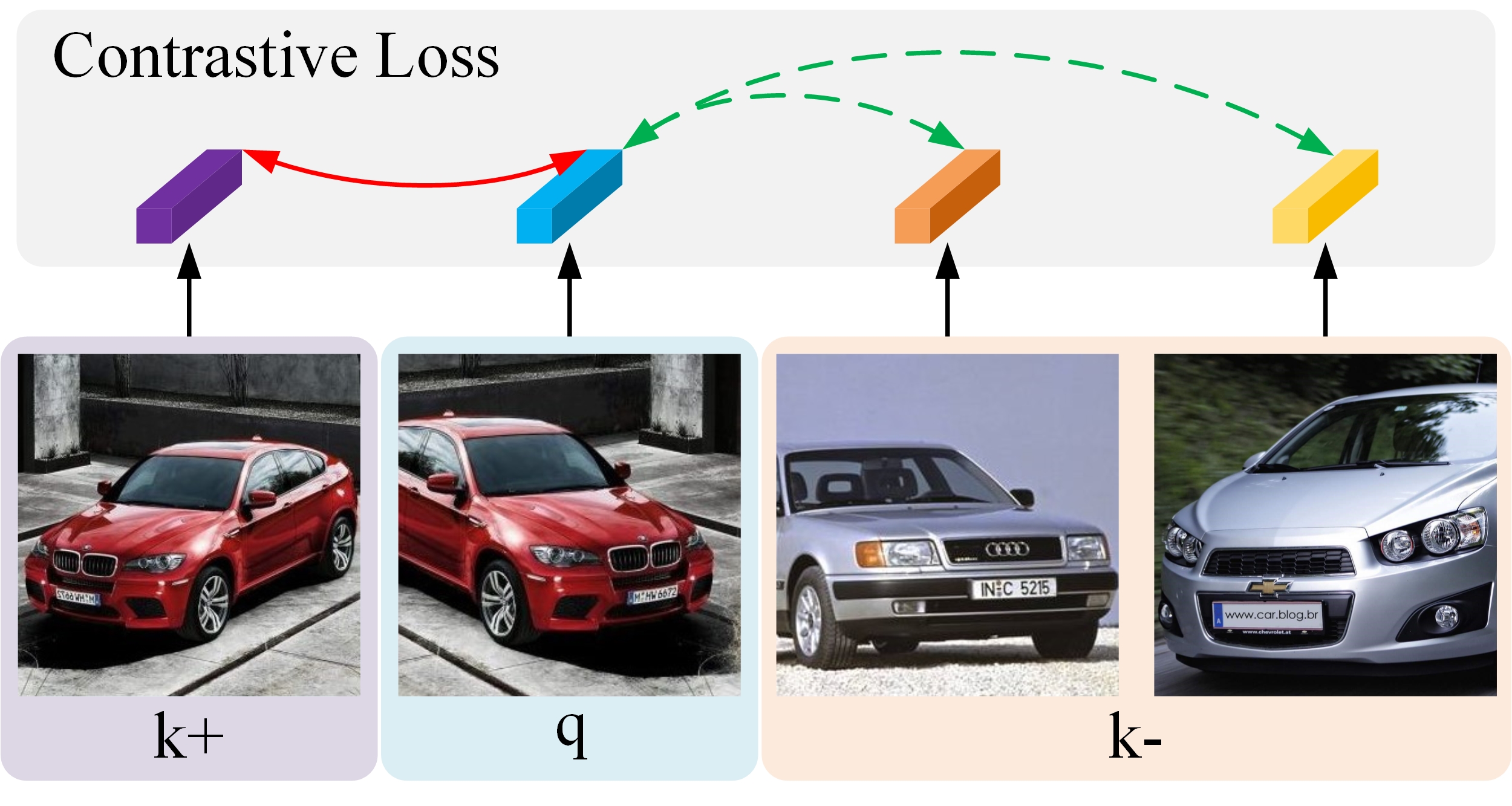}
        }\hfill
    \subfloat[LoDisc]{
        \includegraphics[width=0.48\linewidth]{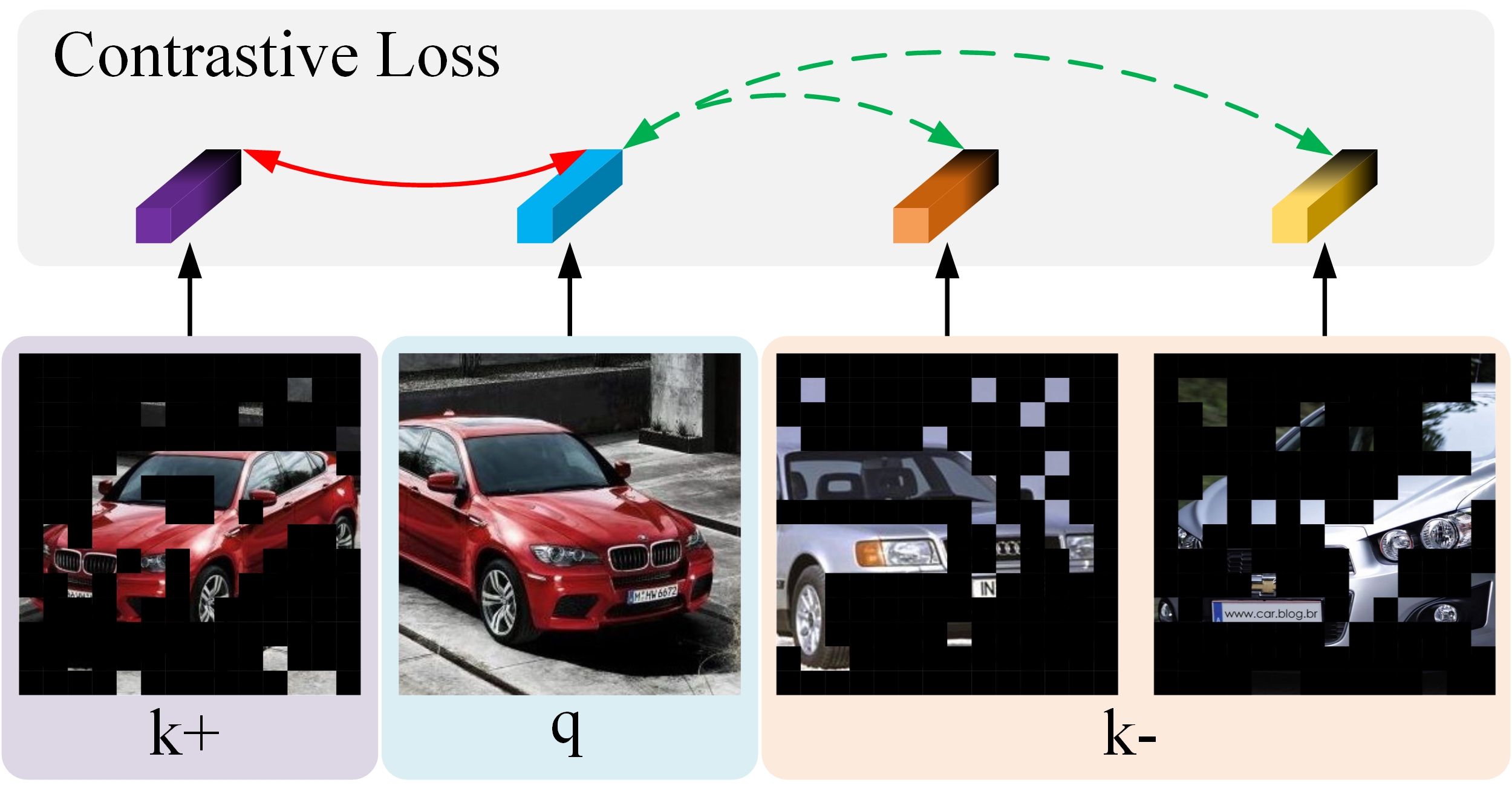}
        }
    \caption{Comparison between two pretext tasks. Positive sample pairs originate from the same sample, such as q and k+. Negative sample pairs originate from different samples, such as q and k-. The two pretext tasks differ in their inputs k. (a) q, k+, and k- are global views from the augmentations of images. (b) Only q is the global view, while both k+ and k- are local regions from the augmentations of images. Contrastive loss maximizes the similarity between positive sample pairs (red solid line) and the dissimilarity between negative sample pairs (green dashed line) to learn discriminative features.}
    \label{fig:2}
\end{figure*}

We present a pure self-supervised global-local fine-grained recognition multi-branch framework by adding a branch for searching fine-grained features at local levels in parallel with the global contrastive learning process, as shown in Fig.~\ref{fig:1}. To explicitly supervise the local branch's focus toward important local regions, an innovative pretext task called local discrimination (LoDisc) is proposed. Compared with InstDisc, our LoDisc uses an augmentation and another augmentation's important local regions from the same image as positive sample pairs, while stipulating the augmentation of this image and local regions in other images' augmentations as negative sample pairs. The similarities of positive sample pairs and dissimilarities of negative sample pairs are subsequently maximized in a representation space via contrastive loss, as shown in Fig.~\ref{fig:2}.

Driven by LoDisc, a local pivotal region collection and selection module and a local discriminative feature learning module are designed. Although important, it remains challenging to explicitly look at relevant local regions in images without additional annotation. To overcome this dilemma, we search for a relationship between the training process and local object regions. Inspired by previous studies \cite{17, 22, 23}, the multi-head attention mechanism \cite{40} in a vision transformer (ViT) \cite{36} can capture the important regions among a sequence of patches. Some supervised works \cite{22, 18, Intro-tcsvt-3} have suggested that the attention weights generated by the attention mechanism in each transformer layer play a crucial role in the process of capturing important patches. The notion of integrating attention weights is natural and applicable in the self-supervised learning process. Therefore, the local pivotal region collection and selection module collects attention weights from all transformer layers to assess the importance of each patch within every image. Then, a simple but effective location-wise mask sampling strategy is developed to selectively cover relatively unimportant patches according to the masking ratio, and the remaining patches are local pivotal regions that are used as inputs for the next module. The local discriminative feature learning module learns fine-grained features from the remaining patches within the constraints of contrastive loss.

Our global-local implementation learns more refined features. The quality of the learned features is evaluated on three popular fine-grained visual recognition benchmarks (FGVC-Aircraft \cite{Aircraft}, Stanford Cars \cite{Cars} and CUB-200-2011\cite{CUB}) and a general object recognition benchmark (Caltech-101 \cite{Caltech}) for linear probing and image retrieval. Extensive experiments demonstrate the effectiveness of the implementation. For example, the proposed method achieves 5.64\% Top-1 accuracy higher than our baseline method (MoCo v3 \cite{MoCov3}) on FGVC-Aircraft and 12.83\% Top-1 accuracy higher than the recent state-of-the-art self-supervised contrastive method designed for FGVR on Stanford Cars. 

The main contributions can be summarized as follows.
\begin{itemize}
\item A novel and pure self-supervised global-local fine-grained visual recognition multi-branch framework is proposed. Moreover, we present an innovative pretext task called local discrimination (LoDisc) to explicitly supervise the local branch focus toward important local regions. The proposed framework not only provides global insights but also delves into local details, enabling the learning of subtle local fine-grained features.
\item We design a local pivotal region collection and selection module to search relevant local regions in the image without additional annotations and develop a location-wise mask sampling strategy to selectively keep some pivotal regions. Then, a local discriminative feature learning module is designed to learn the fine-grained features in pivotal regions. To the best of our knowledge, we are the first to employ local pivotal regions of images in self-supervised contrastive learning.
\item The experimental results on different benchmark data-sets demonstrate that the global-local implementation achieves excellent performance and even outperforms previous state-of-the-art self-supervised fine-grained visual recognition methods. It is also efficacious in general object recognition tasks.
\end{itemize}

\section{Related work}
\subsection{Self-supervised Contrastive Learning}
Self-supervised contrastive learning explores feature representations from unlabeled data via contrastive loss, which measures the similarity of sample pairs in a representation space. Approaches based on self-supervised contrastive learning have been extensively studied in computer vision \cite{DINO, MOBY, VICReg}. Typically, some approaches model image similarity (dissimilarity) from the augmented views of the same (different) sample pairs and pull together similar pairs while pushing apart dissimilar pairs. For example, Chen et al. \cite{SimCLR} employed end-to-end encoders to encode positive (negative) sample pairs and compared them using contrastive cross entropy loss. He et al. \cite {MoCov1, MoCov2, MoCov3} designed an encoder and a momentum encoder to encode positive (negative) sample pairs and compared them via InfoNCE loss. In addition, many approaches model only the image similarity between two or more similar views. For instance, Grill et al. \cite{BYOL} used the representation of an augmented view to predict the representation of another augmented view in the same image. Chen et al. \cite{SimSiam} explored simple Siamese networks for self-supervised visual representation learning. Zbontar et al. \cite{BarlowTwins} proposed a cross-correlation matrix between positive sample pairs as close as possible to the identity matrix. Nonetheless, the core of these self-supervised contrastive learning approaches is that each image possesses its own global distinctiveness, which lacks the utilization of local information. This weakness is not apparent in coarse-grained visual recognition tasks, but it has certain limitations when dealing with the finer levels of granularity within images. 

Inspired by human vision, when confronted with a broad field of observation, the overall perception is typically captured. As the focus narrows to a smaller area, finer details become discernible. Intuitively, local regions are beneficial for facilitating the extraction of additional information. By targeting local regions, Zhou et al. \cite{iBOT} presented a self-supervised teacher-student framework to perform masked patch prediction. Li et al. \cite{esvit} developed region matching for visual representation learning. Zhang et al. \cite{PQCL} explored patch-level contrasts between two views with exact patch correspondence. Assran et al. \cite{JEPA} used a single context block composed of a subset of patches to predict the representations of various target patches in the same image. Local regions are used in the aforementioned works, similar to our work, but the usage and ideas are different. They discovered the semantic correspondence between local regions while we aim to discover discriminative local regions in an image.

\subsection{Contrastive Generative Learning}
Contrastive learning methods typically use the global view as input and focus on global features in the network while often neglecting the internal structure of the view. In contrast, generative learning methods, represented by masked image modeling (MIM), primarily use some masks to disrupt the global context to model the local relationships in an image; these methods typically include an encoder and a decoder in the network. The complementary advantages of these two approaches have led to their combination, resulting in extensive research on contrastive generative learning methods \cite{TPAMI}. Wang et al. \cite{16} introduced a parallel reconstruction branch in the comparison framework. Tao et al. \cite{SiameseIM} learned semantic alignment by predicting the dense representations of an augmented view based on another masked view from the same image. Chen et al. \cite{MixedAE} designed a homologous recognition pretext task that requires each patch to identify homologous patches from mixed samples composed of multiple images. Gupta et al. \cite{SiamMAE} presented learning visual correspondence from randomly sampled pairs of video frames that are asymmetrically masked. The aforementioned methods combine contrastive learning frameworks with mask-based generative learning frameworks, and they typically incorporate at least one decoder in the network for reconstruction or dense prediction tasks. Although our approach also employs a masking strategy similar to that in generative learning, it is not designed for downstream generation tasks, and our network framework does not include a decoder.

\subsection{Fine-grained Visual Recognition}
Fine-grained visual recognition (FGVR) distinguishes subclasses of a given object category with the highly similar global geometry and appearance of the class by often focusing more on the discriminative parts among subclasses. Many supervised learning approaches have achieved great success in locating the discriminative parts of subclasses by region proposal networks and attention mechanisms. Particularly attention-based part localization methods have gained popularity because of the simplicity of training and have demonstrated effectiveness for the FGVR task. Specifically, a set of ViT-based variants \cite{23, 5, 4, 22, 18, Intro-tcsvt-1, CSVT-1} utilize self-attention maps to extract information from images. For example, He et al. \cite{22} exploited ViT inherent attention weights to choose the location part. Zhang et al. \cite{18} further used a multiple-scale pipeline for learning fine-grained features. Ji et al. \cite{Intro-tcsvt-1} also carried out orthogonal multi-grained assembly within the transformer architecture to achieve a more robust fine-grained representation. Sun et al. \cite{CSVT-1} proposed a hierarchical attention network to improve effective attention and facilitate the learning of fine-grained features. However, obtaining sufficiently strong discriminative regions in the global image is often challenging for the model, as self-attention can sometimes focus on unimportant regions, especially during self-supervised learning. Furthermore, some recent methods enhance important information by excluding unimportant regions, such as the background. Chou et al. \cite{backsuppression} utilized supervised classification confidence scores to extract foreground discriminative features and reduce background noise. Zha et al. \cite{CSVT-2} created foreground masks for localization based on supervised activation maps to refine the foreground regions. However, the crucial information provided by supervised labels is not accessible in self-supervised learning.

To bridge the gap between self-supervised and supervised learning for fine-grained object recognition, several innovative works have been proposed. Wu et al. \cite{CVSA} localized foreground objects by cropping and swapping saliency regions of images, which relies on pretrained saliency detection. Zhao et al.\cite{DiLo} improved localization ability by learning invariance through a copy-and-paste technique. Huang et al. \cite{LEWEL} explored the use of alignment maps combined with a coupled projection head to boost the learned representations. Peng et al. \cite{ContrastiveCrop} created a center-suppressed sampling method to increase the variance of global views and demonstrated that semantically aware localization is effective for self-supervised representation learning. Shu et al. \cite{LCR} identified discriminative regions by fitting the gradient-weighted class activation mapping (Grad-CAM) generated from the global contrastive loss of an instance-based method. This implicit restriction in which only global information is used ignores some subtle local differences. Thus, LoDisc is introduced to learn enhanced local discriminative information that is complementary to global discriminative information.

\section{Approach}
In this section, we first briefly review the self-supervised contrastive framework (in Sec. \ref{sec:3.1.1}) and the self-supervised vision transformer contrastive framework (in Sec. \ref{sec:3.1.2}) as preliminary knowledge. We then introduce the proposed local discrimination, which is organized by a local pivotal region collection and selection module and a local discriminative feature learning module, which is critical in our global-local framework (in Sec. \ref{sec:3.2}).
\subsection{Preliminary}
\label{sec:3.1}
\subsubsection{Self-supervised contrastive framework}
\label{sec:3.1.1}
A common practice in self-supervised contrastive learning is to learn not only similarity from two randomly augmented views of an image but also differences from randomly augmented views of different images at a global level through a variant of Siamese networks \cite{35}. The specific explanation of this practice is as follows. 
Given a mini-batch size \(B\) of samples \(I = \left\{ I_{1},I_{2},\ldots,I_{B} \right\}\), each image undergoes data augmentations to generate augmented view1, denoted as \(x\), and augmented view2, denoted as \(x'\). View1 and view2 of all images are encoded by two encoders \(f_q\) and \(f_k\), respectively, to generate feature representations \(\{z_{q}\}\) and \(\{z_{k}\}\). Consider feature representations from two views of the same image as a positive sample pair and feature representations from views of different images as negative sample pairs. 
Accordingly, when the feature representation \(z_q\) of view1 \(x\) from any image behaves like a "query", its positive sample, denoted as \(z_{k+}\), is the feature representation of view2 \(x'\) from the same image. 
Conversely, its negative sample, denoted as \(z_{k-}\), is the feature representation of view2 from a different image. Therefore, there is only one instance of \(z_{k+}\) and \(B-1\) instances of \(z_{k-}\). Set \(\{z_{k}\}\) consists of \(\{z_{k+}\}\) and \(\{z_{k-}\}\). Then, a contrastive loss function is adopted to ensure that the positive sample pair is similar and that the negative sample pairs are dissimilar. The InfoNCE \cite{InfoNCE} format is adopted here:
\begin{equation}
  \mathcal{L}\left( z_{q},z_{k} \right) = -\mathbf{\log}\frac{\exp\left( z_{q} \cdot z_{k+}/\tau \right)}{{\exp\left( {z_{q} \cdot z_{k+}/\tau} \right)} + {\sum\limits_{z_{k-}}{\exp\left( z_{q} \cdot z_{k-}/\tau \right)}}},
\label{eq:1}
\end{equation}    
where \(\tau\) is a temperature hyperparameter that controls the distribution of feature representations. 
\subsubsection{Self-supervised vision transformer contrastive framework}
\label{sec:3.1.2}
In the self-supervised vision transformer contrastive framework, MoCo v3 \cite{MoCov3} coexists samples in the same mini-batch instead of storing negative samples in the memory queue \cite{MoCov1}. Therefore, two different augmented views \(x\) and \(x'\) are separately encoded by \(f_q\) and \(f_k\), resulting in four feature representations, namely, \(z_{q1}\), \(z_{q2}\), \(z_{k1}\), and \(z_{k2}\). The corresponding contrastive loss function is transformed by Eq.~(\ref{eq:2}), which is employed to facilitate global discriminative feature learning. 
\begin{equation}
  \mathcal{L}\left( {z_{q1},z_{q2},z_{k1},z_{k2}} \right) = 2\tau \cdot \mathcal{L}\left( {z_{q1},z_{k2}} \right) + 2\tau \cdot \mathcal{L}\left( {z_{q2},z_{k1}} \right).
\label{eq:2}
\end{equation}

\subsection{Global-local Self-supervised Fine-grained Contrastive Framework}
\label{sec:3.2}

\begin{figure*}[t]
  \centering
  \includegraphics[width=0.98\linewidth]{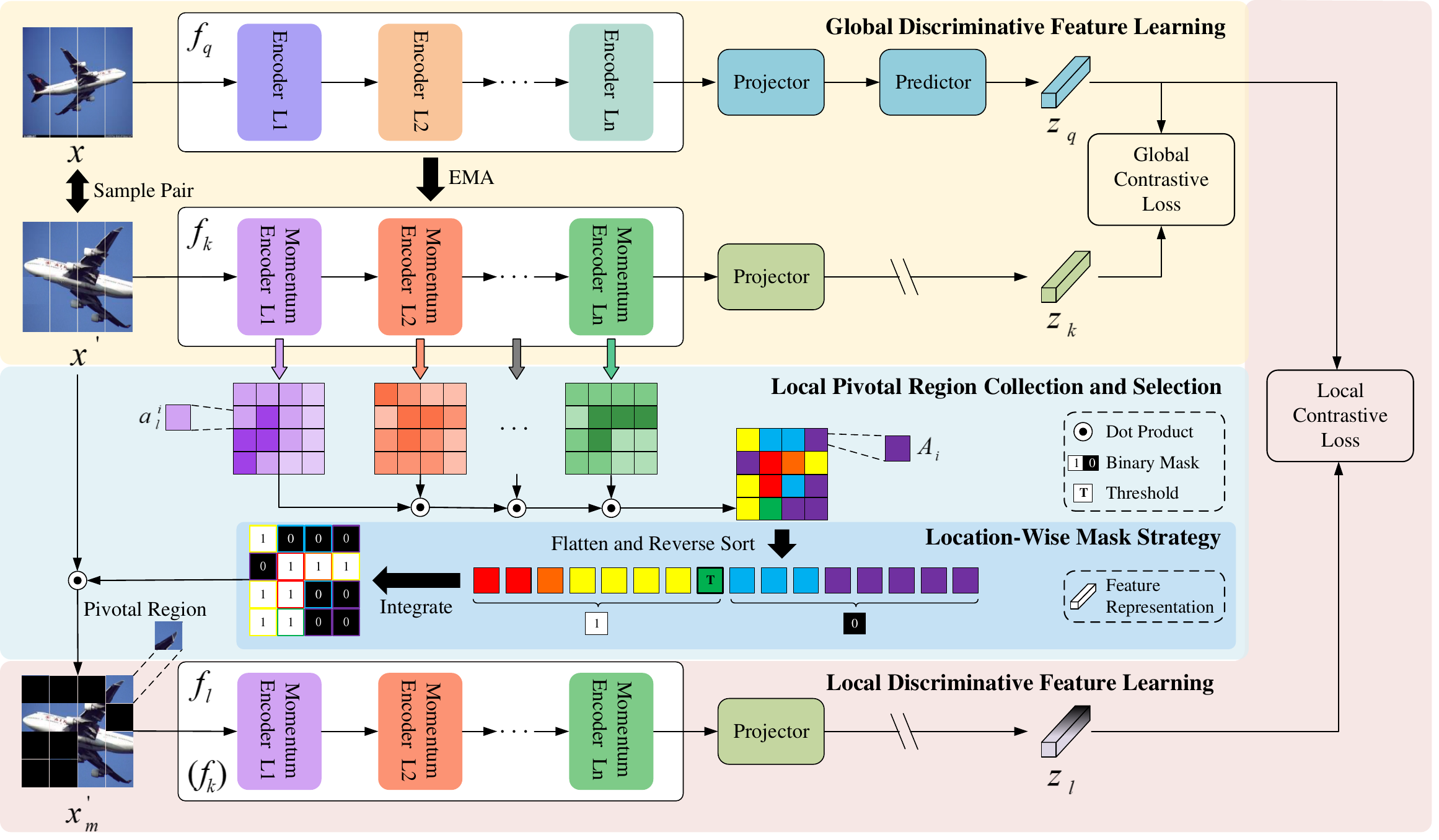}
   \caption{Overview of the proposed method. In the global branches, MoCo v3 is used to learn global coarse-grained discriminative features. To specifically gain local fine-grained features, a pretext task LoDisc is proposed to supervise local branches in learning local fine-grained discriminative features, and it is organized by a local pivotal region collection and selection module and a local discriminative feature learning module. In the first module, the proposed method collects the attention weights of each layer in the momentum encoder, and a location-wise mask sampling strategy is developed to selectively keep local pivotal regions. The second module shares the momentum encoder and learns local discriminative features in pivotal regions. During global coarse-grained learning and local fine-grained learning, the model is optimized by global contrastive loss and local contrastive loss.}
   \label{fig:framework}
\end{figure*}

Unlike common self-supervised contrastive frameworks, which learn feature representations to a global extent, we hope that the self-supervised fine-grained contrastive framework can further focus on important local regions and learn discriminative features from them. This global-local self-supervised fine-grained contrastive framework is illustrated in Fig.~\ref{fig:framework} and elaborated next. First, we introduce a novel pretext task called local discrimination (LoDisc), which is designed to learn local discriminative representations (in Sec. \ref{sec:3.2.1}). Then, we create a local pivotal region collection and selection module to explore and select the local pivotal regions of a given image via a location-wise mask sampling strategy (in Sec. \ref{sec:3.2.2}). A local discriminative feature learning module is subsequently designed to enforce the backbone to learn local fine-grained features (in Sec. \ref{sec:3.2.3}). Finally, the implementation and loss function of the overall global-local framework, including the introduction to its pseudocode, as shown in Algorithm \ref{al1}, are provided in Sec. \ref{sec:3.2.4}.
\subsubsection{Local discrimination}
\label{sec:3.2.1}
Maximizing similar representations of positive sample pairs and dissimilar representations of negative sample pairs is a typical idea in common self-supervised contrastive learning. Importantly, the samples, which are the source of representations, limit the learnable area of the model. Mostly, positive sample pairs are composed of random augmentations of an image, and negative sample pairs are from different images, resulting in global representations. The comparison process of global representations can be regarded as global discrimination. However, during global discrimination, subtle discriminative features important in fine-grained visual recognition tasks are often overlooked. To give attention to the subtle discriminative features, we hypothesize that they can be located in the local pivotal regions of an image. Therefore, the source of representations is changed, which is the core of local discrimination (LoDisc). Specifically, an augmentation and the local pivotal regions from another augmentation of the same image form a positive sample pair, and an augmentation of an image and local pivotal regions of an augmentation of other images form negative sample pairs. Finally, discriminative representations in the local regions are learned through contrastive loss, as shown in Fig.~\ref{fig:2}.

\subsubsection{Local pivotal region collection and selection module}
\label{sec:3.2.2}
Getting local pivotal regions in an image is a prerequisite of LoDisc. However, this is a challenge when there is no additional data annotation. Inspired by previous supervised studies \cite{22, 18}, the attention weights generated by the attention mechanism accumulate gradually as the number of ViT layers increases, allowing the model to capture the important patches we need. We study how local pivotal regions can also be explored by an attention mechanism in the process of self-supervised global discrimination. According to LoDisc, its purpose is to discover the local pivotal regions in each input view of the momentum encoder. The momentum encoder receives a batch of input views. For each randomly augmented input view \({{\mathcal{V}}_{i} \in \mathbb{R}}^{s^{2} \cdot C}\), it is cut into \(N\) non-overlapping and equal-sized patches. Each patch is represented as \(\mathcal{V}_{i}^{j} =\left\lbrack v_{i}^{1},v_{i}^{2},\ldots,v_{i}^{N} \right\rbrack \in \mathbb{R}^{p^{2} \cdot C}\). The patches are then flattened and mapped into a latent \(D\)-dimensional embedding space. The multi-head self-attention mechanism is adopted in each layer of the momentum encoder with \(L\) transformer layers. The mechanism calculates the attention weights between each embedding in the sequence and captures the relationships among them. The attention weights for each layer, directly extracted from the momentum encoder network, are denoted as \( W_{l} \in \mathbb{R}^{B \times H \times (1 + N) \times (1 + N)} \), where \( B \) is the mini-batch size, \( H \) is the number of parallel attention heads, and \( 1 + N \) is the sequence length, which includes one class-token and \( N \) patch-tokens. The average attention weights, represented as \( W_{l}^{avg} \in \mathbb{R}^{B \times (1 + N) \times (1 + N)} \), are calculated by averaging the attention weights across all heads. The class-token aggregates the information from the sequence. Therefore, to identify semantically important regions, we extract the attention weights \( W_{l}^{class} \in \mathbb{R}^{B \times 1 \times N} \) from \( W_{l}^{avg} \) for each layer. The extracted attention weights \( W_{l}^{class} \) consist of a batch of attention weight vectors, with each batch containing \( B \) input views. In the \( l \)-th transformer layer, for each view, there is an attention weight vector associated with the class-token. The elements in this vector represent the relationships between the class-token and each patch-token. This attention weight vector is denoted as
\begin{equation}
a_{l} = \left\lbrack {a_{l}^{1},a_{l}^{2},\ldots,a_{l}^{N}} \right\rbrack \in \mathbb{R}^{1 \times N} ,~l \in \left\{ {1,2,\ldots,L} \right\}, 
\label{eq:new1}
\end{equation}
where \(L\) is the number of layers in the momentum encoder. Therefore, the attention weight vectors from all layers for a given image are collected as
\begin{equation}
{\mathcal{A}} = \bigodot{\!}{}_{_{l = 1}}^{^L}{a}_{l} = \left\lbrack A_{1},A_{2},\ldots,A_{N} \right\rbrack,
\label{eq:3}
\end{equation}
where \(\bigodot\) denotes the Hadamard product.

A simple but effective location-wise mask sampling strategy is subsequently proposed, which further exploits the attention weights. Importantly, in this strategy, the weights are indirectly mapped to the image without modifying the image's feature vectors, which helps mitigate potential concerns. Given two different augmented views of each image from a mini-batch, denoted as \(x\) and \(x'\), when \(x'\) is used as the input view for the momentum encoder, this strategy generates a masked version of \(x'\) for LoDisc, which is denoted as \(x_{m}^{'}\). Similarly, when \(x\) is used as the input view for the momentum encoder, the masked version of \(x\), denoted as \(x_{m}\), is generated. Taking \(x'\) as the input to the momentum encoder to generate \(x_{m}'\) as an example, the process of the strategy is outlined as follows. First, the collection results are reverse-sorted into a list. The patches that correspond to the high attention weights in the sequence are the pivotal regions. Thus, we sequentially sample a subset of patches and pop the last portion of the list with a masking ratio (e.g., 70\%). Specifically, threshold \(A_t\) is the last sampled value in the list; it represents the smallest weight value in the pivotal regions. Second, a list \({\mathcal{M}} \in \mathbb{R}^{1 \times N}\) is created to store the mask values, where \(M_i\in \left\{ {0,1} \right\}\). The attention weights in \( {\mathcal{A}} \) are compared with the threshold \(A_t\) sequentially. If the attention weight of a patch in the image is higher than the threshold, then this patch considers a pivotal region and records a mask value of 1 in \({\mathcal{M}}\) at the corresponding position. Otherwise, a mask value of 0 is recorded in \({\mathcal{M}}\), indicating that the patch should be masked. This process is denoted as
\begin{equation}
\begin{split}
{M}_{i} = \left\{ \begin{matrix}{~~0,~~~~~A_{i} < A_{t}} \\
{~~1,~~~~~A_{i} \geq A_{t}}\end{matrix} \right..
\label{eq:also-important}
\end{split}
\end{equation}
Next, the list \({\mathcal{M}} \in \mathbb{R}^{1 \times N}\) is transformed into a matrix \( \textbf{G} \in \mathbb{R}^{\sqrt{N} \times \sqrt{N}}\), where each value \(\textbf{G}_{i,j}\) is expanded into a pixel matrix of patch size \(\textbf{P} \in \mathbb{R}^{p \times p}\). Therefore, we obtain a mask-pixel matrix \({\textbf{MP} \in \mathbb{R}}^{s \times s} \) with mask values that match the size of \(x'\), where \(\textbf{MP}_{i,j} \in \{0, 1\}\). \(\textbf{MP}\) is replicated along a specified dimension to ensure that its shape matches the number of channels in \(x'\). Finally, \(x_{m}^{'}\) is created by \(\textbf{MP} \bigodot x'\).

\subsubsection{Local discriminative feature learning module}
\label{sec:3.2.3}
Based on the method described above, we can obtain an augmented view \(x\) of an image along with its masked version \(x_{m}\). Similarly, we can derive another augmented view \(x'\) from the same image and generate a masked version \(x_{m}^{'}\) that corresponds to \(x'\). In alignment with global discriminative feature learning, \(x\) and \(x'\) are encoded by an encoder \(f_q\) to generate feature representations \(z_{q1}\) and \(z_{q2}\). Unlike global discriminative feature learning, the momentum encoder generates features using only the masked version of the view as input, so \(x_{m}\) and \(x_{m}^{'}\) are encoded by a shared momentum encoder \(f_l\) (which is equivalent to \(f_k\)) to generate feature representations \(z_{l1}\) and \(z_{l2}\). The same contrastive loss with Eq.~(\ref{eq:2}) is adopted to explicitly supervise the encoder focus toward local pivotal regions limited by the input regions of \(x_{m}\) and \(x_{m}^{'}\).

Similar to the common self-supervised vision transformer contrastive framework \cite{MoCov3}, \(f_q\) consists of a ViT backbone, a projection head \cite{SimCLR} and a prediction head \cite{BYOL}. \(f_l\) is shared with the momentum encoder \(f_k\) of global discrimination, which is composed of a ViT backbone and a projection head. At the inference time, the output of \(f_q\) is adopted.

\begin{algorithm}[!ht]
\caption{Pseudocode of Global-Local Self-Supervised Fine-Grained Learning on MoCov3, PyTorch-style.}
\label{al1}
\begin{algorithmic}[1] % 控制是否有序号
    \STATE \# \(f_q\): encoder: backbone + projector + predictor
    \STATE \# \(f_k\): momentum encoder: backbone + projector
    \STATE \# \(\mathbf{\beta}\): momentum coefficient
    \STATE \# \(\tau\): temperature
    \STATE \# \(I\): a mini-batch \(I\) with \(B\) samples
    \STATE
    \FOR{\(I\) in dataloader:}
    \STATE \# Sec. \ref{sec:3.1}
    \STATE $x, x^{'} = aug(I), aug(I)$
    \STATE $z_{q1},z_{q2} = f_{q}(x), f_{q}(x^{'})$
    \STATE $z_{k1},z_{k2} = f_{k}(x), f_{k}(x^{'})$
    \STATE $\mathcal{L}_G = CL(z_{q1}, z_{k2}) + CL(z_{q2}, z_{k1})$ \# global\_loss
    \STATE 
    \STATE \# Sec. \ref{sec:3.2.2}
    \STATE $att_{1}, att_{2} = get\_ att(x, f_{k}), get\_ att(x^{'}, f_{k})$ \# Extract attention map from each layer of momentum encoder
    \STATE $f\_ att_{1}, f\_ att_{2} = fuse\_ att(att_{1}), fuse\_ att(att_{2})$ \# Extract and aggregate attention map between class-token and each patch-token
    \STATE $mask_{1}, mask_{2} = mask\_ att(f\_ att_{1}), mask\_ att(f\_ att_{2})$ \# Get masking map
    \STATE $x_{m}, x_{m}^{'} = app\_ mask(x, mask_{1}), app\_ mask(x^{'}, mask_{2})$ \# Apply masking map to view
    \STATE \# Sec. \ref{sec:3.2.3} 
    \STATE $z_{l1},z_{l2} = f_{k}(x_{m}), f_{k}(x_{m}^{'})$
    \STATE $\mathcal{L}_L= CL(z_{q1}, z_{l2})+ CL(z_{q2}, z_{l1})$ \# local\_loss
    \STATE \# Sec. \ref{sec:3.2.4}
    \STATE $\mathcal{L}= \mathcal{L}_G+\mathcal{L}_L$
    \STATE $\mathcal{L}.backward()$
    \STATE 
    \STATE $update(f_{q})$
    \STATE $f_{k}=\beta*f_{k}+(1-\beta)*f_{q}$
    \ENDFOR
    \STATE  
    \FUNCTION{cl}{q, k}
    \STATE \quad $logits=mm(q, k.t())$
    \STATE \quad $labels=range(B)$
    \STATE \quad $loss=CrossEntropyLoss(logits/\tau, labels)$
    \STATE \quad \textbf{return} $2*\tau*loss$
    \ENDFUNCTION
\end{algorithmic}
\end{algorithm}

\subsubsection{Global-local self-supervised fine-grained learning}
\label{sec:3.2.4}
To encourage the model to learn valuable features from both the global and pivotal local perspectives, LoDisc combines with global discrimination, and the overall global-local framework is implemented by multiple parallel branches (as shown in Fig.~\ref{fig:1}) and optimized with global contrastive loss and local contrastive loss. The global contrastive loss measures the similarity between the representations of \(x\) and \(x'\); the local contrastive loss measures the similarity between the representations of \(x\) and \(x_{m}^{'}\) and between the representations of \(x'\) and \(x_{m}\); and they simultaneously learn features. The overall loss function is calculated as
\begin{equation}
    \mathcal{L} = \mathcal{L}\left( {z_{q1},z_{q2},z_{k1},z_{k2}} \right) + \mathcal{L}\left( {z_{q1},z_{q2},z_{l1},z_{l2}} \right).
\label{eq:also-important}
\end{equation}

\section{Experiments}
In this section, the aim is to demonstrate the representation learning ability of the proposed global-local self-supervised fine-grained contrastive framework. We first provide a short description of the datasets and experimental settings (in Sec.~\ref{sec:4.1}). Afterwards, we conduct experiments involving the proposed approach by comparing it with self-supervised contrastive methods (in Sec.~\ref{sec:4.2}), self-supervised localization methods (in Sec.~\ref{sec:4.add}), and a self-supervised region-level method (in Sec.~\ref{sec:4.add2})), and we perform key parameter studies (in Sec.~\ref{sec:4.3}) and ablation studies (in Sec.~\ref{sec:4.4}). We apply the global-local framework to general object recognition (in Sec.~\ref{sec:4.5}), followed by a visual analysis as the final step (in Sec.~\ref{sec:4.6}).

\subsection{Datasets and Experimental Settings}
\label{sec:4.1}
\subsubsection{Datasets}
We evaluate the performance of our global-local framework on three fine-grained visual datasets (FGVC-Aircraft \cite{Aircraft}, Stanford Cars \cite{Cars} and CUB-200-2011\cite{CUB}) and a generic visual dataset (Caltech-101 \cite{Caltech}). The detailed descriptions of these four datasets are as follows.

\begin{itemize}
\item \textbf{FGVC-Aircraft} \cite{Aircraft} comprises 10,000 images from 100 categories of aircraft, e.g., Cessna-560 and Fokker-100, with a split of 6,667 training images and 3,333 testing images. 
\item \textbf{Stanford Cars} \cite{Cars} comprises 16,185 images from 196 categories of cars, e.g., Audi-A5 and BMW-X3, with a split of 8,144 training images and 8,041 testing images. 
\item \textbf{CUB-200-2011} \cite{CUB} comprises 11,788 images from 200 categories of birds, e.g., Rusty Blackbird and Brewer Blackbird, with a split of 5,994 training images and 5,794 testing images. 
\item \textbf{Caltech-101} \cite{Caltech} comprises 9,145 images from 101 object categories and a background category, e.g., lotus and ant, with a split of 6,822 training images and 2,323 testing images.
\end{itemize}

\begin{table*}
\caption{Comparison of different self-supervised contrastive methods in terms of Top-1 accuracy (\%) on FGVC-Aircraft, Stanford Cars, and CUB-200-2011, using a batch size of 32; "\textnormal{Param.}" is the total number of parameters (without counting the projection head), “\textnormal{im/s}” is the inference time on a NVIDIA V100 GPU with 128 samples per forward; "T\(_{100}\)" is the running time on 4 NVIDIA V100 GPUs with 100 epochs and "\textnormal{Mem.}" is the peak memory per GPU on the CUB-200-2011 dataset.}
\centering
\label{tab:1}
\begin{tabular}{c|c|c|c|c|c|c|c|c|c|c}
\toprule
\multicolumn{1}{c|}{\multirow{2}{*}{Method}} & \multicolumn{1}{c|}{\multirow{2}{*}{Reference}} & \multicolumn{1}{c|}{\multirow{2}{*}{Backbone}} & \multicolumn{1}{c|}{\multirow{2}{*}{Param.}} & \multicolumn{1}{c|}{\multirow{2}{*}{\hspace{0.5em}im/s\hspace{0.5em}}}& \multicolumn{1}{c|}{\multirow{2}{*}{Batch Size}} & \multicolumn{3}{c|}{\multirow{1}{*}{Top-1 Accuracy (\%)}} & \multicolumn{2}{c}{\multirow{1}{*}{CUB}} \\ 
\cmidrule(r){7-11}& & & & & & Aircraft & \hspace{1em}Cars\hspace{1em}  & \hspace{1em}CUB\hspace{1em} & \hspace{0.5em}T\(_{100}\)\hspace{0.5em} & \hspace{0.5em}Mem.\hspace{0.5em} \\
\midrule
MoCo v2 \cite{MoCov2} & CVPR 2020 & ResNet-50 & 23M & 1237 & 32 & 42.51 & 52.61   & 68.03 & 2.0h & 6.1G\\
SimCLR \cite{SimCLR} & ICML 2020 & ResNet-50 & 23M & 1237 & 32  & 40.56 & 44.31   & 33.49 & 2.5h & 7.1G\\
BYOL \cite{BYOL} & NeurIPS 2020 & ResNet-50 & 23M & 1237 & 32  & 34.90 & 43.66   & 36.64 & 4.0h & 7.4G\\
DINO \cite{DINO} & ICCV 2021 & ResNet-50 & 23M & 1237 & 32   & 8.52 & 9.27 & 12.37 & 1.5h & 4.9G\\
SimSiam \cite{SimSiam} & CVPR 2021 & ResNet-50 & 23M & 1237 & 32  & 41.59 & 56.87 & 35.82& 2.0h & 4.4G\\  
BarlowTwins \cite{BarlowTwins} & ICML 2021 & ResNet-50 & 23M & 1237 & 32  & 28.35 & 23.34 & 28.58 & 1.5h & 7.0G\\
VICReg \cite{VICReg} & arXiv 2021 & ResNet-50 & 23M & 1237 & 32  & 29.97 & 19.29   & 30.37 & 1.0h & 7.0G\\
MoCo v3 \cite{MoCov3} & ICCV 2021 & ViT-S & 21M & 1007 & 32 & 47.92 &  51.36  & 60.43 & 3.0h & 6.5G\\
IBOT \cite{iBOT} & ICLR 2022 & ViT-B & 85M & 312  & 32 & 27.87 & 21.31  & 23.89 & — & 8.6G\\
LCR \cite{LCR} & CVPR 2023 & ResNet-50 & 23M & 1237  & 32 & 46.92 & 56.90   & 70.43 & 2.0h & 6.1G\\
\midrule
\multicolumn{1}{c|}{\multirow{2}{*}{Ours}}  & \multicolumn{1}{c|}{\multirow{2}{*}{2024}} & ViT-S  & 21M & 1007 & 32 & 54.52 &  58.15  & 61.82 & 4.0h & 7.0G\\
 &  & ViT-B  & 85M & 312 & 32 & \textbf{63.40} & \textbf{69.73}  & \textbf{72.66} & 7.0h & 9.0G\\
\bottomrule
\end{tabular}
\end{table*}

\begin{table}
\caption{Comparison of the different self-supervised contrastive methods in terms of Top-1 accuracy (\%) on FGVC-Aircraft, Stanford Cars, and CUB-200-2011 using a batch size of 128.}
\centering
\label{tab:add1}
\begin{tabular}{c|c|c|c|c}
\toprule
Method & Backbone & Aircraft & \hspace{1em}Cars\hspace{1em} & \hspace{1em}CUB\hspace{1em}\\
\midrule
MoCo v2 \cite{MoCov2} & ResNet-50 & 52.54 & 58.43 & 68.30\\
SimCLR \cite{SimCLR} & ResNet-50 & 45.22 & 49.41 & 38.39 \\
BYOL \cite{BYOL} & ResNet-50 & 37.62 & 45.21 & 39.27 \\
DINO \cite{DINO} & ResNet-50 & 12.93 & 10.51 &  16.66 \\
SimSiam \cite{SimSiam} & ResNet-50 & 43.06 & 58.89 & 39.97 \\  
BarlowTwins \cite{BarlowTwins} & ResNet-50 & 34.77 & 31.91 & 33.45 \\
VICReg \cite{VICReg} & ResNet-50 & 36.00 & 30.80 & 37.78 \\
MoCo v3 \cite{MoCov3} & ViT-S & 51.19 & 59.46 & 66.95 \\
EsViT \cite{esvit} & Swin-T &  55.18 & 59.12 & 70.54 \\
LCR \cite{LCR} & ResNet-50 & 55.87 & 60.75 & 71.31 \\
\midrule
\multicolumn{1}{c|}{\multirow{2}{*}{Ours}} & ViT-S & 60.52  &  65.47  & 67.92 \\
 & ViT-B & \textbf{62.17} & \textbf{69.72}  & \textbf{73.23}\\
\bottomrule
\end{tabular}
\end{table}

\subsubsection{Implementation details}
We adopt ViT-S\_16 and ViT-B\_16 \cite{36} as the backbone of the encoder in the proposed framework and load weights from the official ViT-S\_16 and ViT-B\_16 models pretrained on ImageNet-1K. The images from all of the datasets are resized to 224 \(\times\) 224 \(\times\) 3 and cut into patches of size 16 \(\times\) 16. In addition, we closely follow the designs in MoCo v3 \cite{MoCov3}. The momentum value and hidden layers of both MLPs are set similarly to MoCo v3, i.e., 0.99 and 4096-d, respectively. However, when ViT-S\_16 serves as the backbone network, the output layers of both MLPs are changed to 384-d; when ViT-B\_16 serves as the backbone network, the output layers of both MLPs are changed to 768-d. We set the mini-batch sizes as 32 and 128 and use the AdamW optimizer with a weight decay of 0.5 and a cosine annealing schedule. 100 epochs are used to train the feature extractor. After training, following common practice, the MLP heads are removed, and a linear classifier is trained on the frozen features for linear probing. We set the mini-batch size to 256 and use an SGD optimizer with a weight decay of 0, a cosine annealing schedule and a learning rate of 0.5. 100 epochs are used to train the linear classifier. We conducted experiments using Nvidia Tesla V100 GPUs. 

\subsubsection{Evaluation protocols}
As stated in \cite{DINO,LCR}, universal protocols for self-supervised learning include linear probing and image retrieval. In linear probing, a supervised linear classifier is trained on frozen features from a feature extractor that is trained using self-supervised learning. Therefore, the accuracy of the linear classifier reflects the quality of the features learned by the model during self-supervised learning. The Top-1 accuracy and Top-5 accuracy are metrics used to evaluate classification performance. Concurrently, the Top-1 accuracy is a standard metric for fine-grained visual recognition. In image retrieval, images in the same category are searched based on the similarity of the features from a given query image, which is analogous to performing the nearest neighbor classification task. Higher-quality features from the self-supervised feature extractor, which captures the most relevant information about the images, are more likely to yield accurate and relevant search results. Accordingly, the quality of the features is directly related to the retrieval accuracy. Rank-1, Rank-5 and mAP are used to evaluate the retrieval performance.

\subsection{Comparison with Self-supervised Contrastive Methods}
\label{sec:4.2}
We compare the proposed global-local framework with other popular self-supervised contrastive frameworks on the abovementioned fine-grained datasets. For fairness, all of the following models are uniformly set with mini-batch sizes of 32 and 128 and initialized with ImageNet-1K-trained weights. Here, the masking ratio in our model is set to 70\%. 

Table~\ref{tab:1} shows the classification performance on FGVC-Aircraft, Stanford Cars, and CUB-200-2011 using a mini-batch size of 32. The results of \cite{DINO, SimCLR, BYOL, SimSiam, MoCov2, BarlowTwins, VICReg, LCR} on the datasets come from \cite{LCR}. Specifically, when ViT-S is used as the backbone network, the proposed method achieves Top-1 accuracy of 54.52\%, 58.15\% and 61.82\% on the FGVC-Aircraft, Stanford Cars and CUB-200-2011 datasets, respectively. This demonstrates good classification performance on most datasets and exceeds the state-of-the-art methods by 6.60\% in the Top-1 accuracy for FGVC-Aircraft and 1.25\% for Stanford Cars. Compared with the base self-supervised vision transformer contrastive method MoCo v3, we obtain a 6.60\%, 6.79\% and 1.39\% improvement on the FGVC-Aircraft, Stanford Cars and CUB-200-2011 datasets, respectively. When ViT-B is used as the backbone network, our global-local framework achieves Top-1 accuracy of 63.40\%, 69.73\% and 72.66\% on the FGVC-Aircraft, Stanford Cars and CUB-200-2011 datasets, respectively, which outperform MoCo v3 and all CNN-based methods.

Additionally, a comprehensive comparison is provided in terms of the total number of parameters, inference time, running time, and peak memory in Table~\ref{tab:1}. With respect to the running time and peak memory, our method shows slightly lower training efficiency than the other methods show, primarily because of its use of ViT as the backbone. The model has a more complex structure than the CNN-based methods have, which requires more computational steps to converge and leads to a longer training time and increased memory consumption. Other important metrics that reflect model efficiency include the total number of parameters and inference time. From the experimental results, with fewer parameters and a reduced inference time, our method achieves optimal Top-1 performance on most datasets; with more parameters and an increased inference time, it achieves the best performance on all datasets.

Table~\ref{tab:add1} shows the classification performance on FGVC-Aircraft, Stanford Cars, and CUB-200-2011 using a mini-batch size of 128. The experimental results indicate that the proposed method achieves excellent classification performance on different datasets, resulting in significant improvements. Specifically, when ViT-S is used as the backbone network, which has a comparable number of parameters to LCR, our method outperforms the state-of-the-art method LCR by 4.65\% and 4.72\% in terms of the Top-1 accuracy for the FGVC-Aircraft and Stanford Cars datasets, respectively. For the CUB-200-2011 dataset, our method's Top-1 accuracy is slightly inferior to that of LCR but still higher than that of most other methods. When ViT-B is used as the backbone network, our method further improves the results and achieves the best accuracy. These results highlight the effectiveness of our method in capturing subtle inter-class differences.

\begin{table*}
\caption{Comparison of the different self-supervised localization methods in terms of Top-1 accuracy (\%) and Rank-1 accuracy (\%) on FGVC-Aircraft, Stanford Cars, and CUB-200-2011 using a batch size of 128.}
\centering
\label{tab:add2}
\begin{tabular}{c|c|c|c|c|c|c|c|c|c}
\toprule
\multicolumn{1}{c|}{\multirow{2}{*}{Method}} & \multicolumn{1}{c|}{\multirow{2}{*}{Reference}} & \multicolumn{1}{c|}{\multirow{2}{*}{Backbone}} & \multicolumn{1}{c|}{\multirow{2}{*}{Batch Size}} & \multicolumn{3}{c|}{\multirow{1}{*}{Classification}} & \multicolumn{3}{c}{\multirow{1}{*}{Retrieval}} \\ 
\cmidrule(r){5-10} & & & & Aircraft & \hspace{1em}Cars\hspace{1em}  & \hspace{1em}CUB\hspace{1em} & Aircraft & \hspace{1em}Cars\hspace{1em}  & \hspace{1em}CUB\hspace{1em}  \\
\midrule
DiLo \cite{DiLo} & AAAI 2021 & ResNet-50 & 128 & --  &  --  & 64.14 &  --   & -- &  -- \\
CVSA \cite{CVSA} & arXiv 2021 & ResNet-50 & 128 &  -- &  --  & 65.02 &   --  & -- & --  \\
LEWEL \cite{LEWEL} & CVPR 2022 & ResNet-50 & 128 & 54.33  &  59.02 & 69.27 & 20.67  & 12.01  & 19.23 \\
ContrastiveCrop \cite{ContrastiveCrop}  & CVPR 2022 & ResNet-50 & 128 &  54.40 & 61.66 & 68.82 & 20.88  & 13.61 & 18.71  \\
SAM-SSL \cite{SAM-SSL} & ECCV 2022& ResNet-50 & 128 & 52.97  & 58.49  & 68.59 & 21.72 & 14.26 & 18.38 \\
EsViT \cite{esvit}  & ICLR 2022 & Swin-T & 128 & 55.18  &  61.98  & 70.54 &  27.06 & 31.95 &  43.48 \\
\midrule
\multicolumn{1}{c|}{\multirow{2}{*}{Ours}}  & \multicolumn{1}{c|}{\multirow{2}{*}{2024}} & ViT-S  & 128 &  60.52 &  65.47  & 67.92 &   36.63  & 32.82 &  40.46 \\
 & & ViT-B  & 128 & \textbf{62.17} & \textbf{69.72}  &  \textbf{73.23}  & \textbf{41.49} &   \textbf{41.55}  &  \textbf{45.89} \\
\bottomrule
\end{tabular}
\end{table*}

\begin{table*}
\centering
  \caption{Comparison of the self-supervised region-level methods in terms of Top-1 accuracy (\%) on FGVC-Aircraft, Stanford Cars, and CUB-200-2011 using a batch size of 128.}
  \label{tab:add4}
  \begin{tabular}{c|c|c|c|cc|ccc}
    \toprule
    \multicolumn{1}{c|}{\multirow{2}{*}{Dataset}} & \multicolumn{1}{c|}{\multirow{2}{*}{Method}} & \multicolumn{1}{c|}{\multirow{2}{*}{Backbone}} & \multicolumn{1}{c|}{\multirow{2}{*}{Param.}} & \multicolumn{2}{c|}{\multirow{1}{*}{Classification}} & \multicolumn{3}{c}{\multirow{1}{*}{Retrieval}} \\
    \cmidrule(r){5-9} & & & & Top-1 & Top-5 & Rank-1 & Rank-5 & mAP \\
    \midrule
     \multicolumn{1}{c|}{\multirow{2}{*}{Aircraft}} & EsViT & Swin-T & 28M & 55.18 & 85.36 & 27.06 & 53.02 & 13.58\\
     & Ours & ViT-S & 21M & 60.52 & 88.45 & 36.63 & 63.37 & 18.59\\
     \midrule
     \multicolumn{1}{c|}{\multirow{2}{*}{Cars}} & EsViT & Swin-T & 28M & 59.12 & 87.84 & 31.95 & 58.40 & 9.22\\
     & Ours & ViT-S & 21M & 65.47 & 88.61 & 32.82 & 57.03 & 9.25\\
     \midrule
     \multicolumn{1}{c|}{\multirow{2}{*}{CUB}} & EsViT & Swin-T & 28M & 70.54 & 92.92 & 43.48 & 73.08 & 24.55\\
     & Ours & ViT-S & 21M & 67.92 & 90.04 & 40.46 & 67.24 & 21.25\\
    \bottomrule
  \end{tabular}
\end{table*}

\subsection{Comparison with Self-supervised Localization Methods}
\label{sec:4.add}
Table~\ref{tab:add2} shows the classification and retrieval performance on FGVC-Aircraft, Stanford Cars, and CUB-200-2011 using a mini-batch size of 128. The results of \cite{DiLo, CVSA, LEWEL, ContrastiveCrop, SAM-SSL} on the datasets come from \cite{LCR}. The proposed method yields impressive experimental results on multiple datasets. When ViT-S is used as the backbone network, it outperforms the state-of-the-art methods by 5.34\% and 3.49\% in Top-1 accuracy for the FGVC-Aircraft and Stanford Cars datasets, respectively, and surpasses them by 9.57\% and 0.87\% in retrieval accuracy on the same datasets. For CUB-200-2011, its Top-1 and Rank-1 values are slightly inferior to those of EsViT, but they are still significantly higher than those of the other methods. When ViT-B is used as the backbone network, the proposed method achieves the best classification and retrieval performance and significantly surpasses the second-best method by 6.99\%, 7.74\%, and 2.69\% in Top-1 accuracy across the FGVC-Aircraft, Stanford Cars and CUB-200-2011 datasetss, respectively; additionally, it outperforms the second-best method by 14.43\%, 9.60\%, and 2.41\% in Rank-1 accuracy on the same three datasets. 

\begin{table}
  \caption{Performance of the different masking ratios on FGVC-Aircraft and Stanford Cars using a mini-batch size of 32 and the ViT-B backbone network.}
  \label{tab:3}
  \centering
  \begin{tabular}{c|c|ccc}
    \toprule
    \multicolumn{1}{c|}{\multirow{2}{*}{Dataset}} & \multicolumn{1}{c|}{\multirow{2}{*}{Masking ratio}} & \multicolumn{3}{c}{\multirow{1}{*}{Retrieval}} \\
    \cmidrule(r){3-5} & & Rank-1 & Rank-5 & mAP \\
    \midrule
    \multicolumn{1}{c|}{\multirow{3}{*}{Aircraft}} & 60\% & 36.06 & 64.57 & 21.08\\
    & 70\% & 38.52 & \textbf{66.82} & \textbf{21.11}\\
    & 80\% & \textbf{38.91} & 66.79 & 20.50\\
    \midrule
    \multicolumn{1}{c|}{\multirow{3}{*}{Cars}} 
    & 60\% &  38.34 & \textbf{65.69}  & \textbf{11.87} \\
    & 70\% &  \textbf{38.37} & 64.45  & 11.29 \\
    & 80\% &  32.57 & 56.74  &  9.05\\
    \bottomrule
  \end{tabular}
\end{table}

\begin{table*}
  \caption{The classification and retrieval performance of the proposed method are evaluated on FGVC-Aircraft, Stanford Cars and CUB-200-2011 using ViT-S as the backbone network. We report the Top-1 and Top-5 accuracy(\%) on the classification task and the Rank-1, Rank-5, and mAP accuracy(\%) on the retrieval task. "Moco v3 (g)" represents a self-supervised contrastive method that contains only global branches. "Ours (l)" represents our proposed method that contains only local branches, and its pretext task is our proposed LoDisc. "Ours (g+l)" represents the proposed method that contains both global and local branches.}
  \label{tab:add3}
  \centering
  \begin{tabular}{c|c|c|cc|ccc}
    \toprule
    \multicolumn{1}{c|}{\multirow{2}{*}{Dataset}} & \multicolumn{1}{c|}{\multirow{2}{*}{Method}} & \multicolumn{1}{c|}{\multirow{2}{*}{Batch size}} & \multicolumn{2}{c|}{\multirow{1}{*}{Classification}} & \multicolumn{3}{c}{\multirow{1}{*}{Retrieval}} \\
    \cmidrule(r){4-8} & & & Top-1 & Top-5 & Rank-1 & Rank-5 & mAP \\
    \midrule
     \multicolumn{1}{c|}{\multirow{3}{*}{Aircraft}} & 
   MoCo v3 (g) & 32/128 & 47.92/51.19 & 79.93/83.56 & 23.13/26.82 & 45.42/51.43 & 10.46/13.75\\
    & Ours (l) & 32/128 & 53.35/60.16 & \textbf{84.16}/86.92 & 33.33/36.57 & 59.32/63.31 & 16.27/18.13\\
    & Ours (g+l) & 32/128 & \textbf{54.52/60.52} & 83.29/\textbf{88.45} & \textbf{33.84/36.63} & \textbf{60.40/63.37} & \textbf{17.26/18.59}\\
     \cmidrule(r){1-8}
    \multicolumn{1}{c|}{\multirow{3}{*}{Cars}} & 
   MoCo v3 (g) & 32/128 & 51.36/59.46  & 80.20/85.24 & 21.49/25.16 & 43.13/47.08 & 5.50/5.91 \\
    & Ours (l) & 32/128 & 50.58/54.63 & 79.12/81.63 & 18.49/22.78 & 38.25/45.13 & 4.75/5.28 \\
    & Ours (g+l) & 32/128 & \textbf{58.15/65.47} & \textbf{84.44/88.61} & \textbf{23.53/32.82} & \textbf{44.47/57.03} & \textbf{5.88/9.25} \\
     \cmidrule(r){1-8}
     \multicolumn{1}{c|}{\multirow{3}{*}{CUB}} & 
   MoCo v3 (g) & 32/128 & 60.43/66.95 & 85.26/89.56 & 31.10/36.14 & 58.58/64.88 & 12.44/15.56 \\
    & Ours (l) & 32/128 & 61.37/67.79 & 84.88/88.99 & 34.88/39.21 & 61.55/65.07 & 16.57/19.74 \\
    & Ours (g+l) & 32/128 & \textbf{61.82/67.92} & \textbf{86.00/90.04} & \textbf{36.21/40.46} & \textbf{61.93/67.24} & \textbf{17.84/21.25} \\  
    \bottomrule
  \end{tabular}
\end{table*}

\begin{table*}
  \caption{The classification and retrieval performance of the proposed method are evaluated on FGVC-Aircraft, Stanford Cars and CUB-200-2011 using ViT-B as the backbone network.}
  \label{tab:2}
  \centering
  \begin{tabular}{c|c|c|cc|ccc}
    \toprule
    \multicolumn{1}{c|}{\multirow{2}{*}{Dataset}} & \multicolumn{1}{c|}{\multirow{2}{*}{Method}} & \multicolumn{1}{c|}{\multirow{2}{*}{Batch size}} & \multicolumn{2}{c|}{\multirow{1}{*}{Classification}} & \multicolumn{3}{c}{\multirow{1}{*}{Retrieval}} \\
    \cmidrule(r){4-8} & & & Top-1 & Top-5 & Rank-1 & Rank-5 & mAP \\
    \midrule
     \multicolumn{1}{c|}{\multirow{3}{*}{Aircraft}} & 
   MoCo v3 (g) & 32/128 & 57.76/58.84 & 87.16/88.15  & 32.37/33.75  & 58.03/60.19  & 16.72/19.01 \\
    & Ours (l) & 32/128 & 59.08/61.81  & 87.49/88.00  & 34.80/40.23 & 61.69/67.45  & 16.85/21.13 \\
    & Ours (g+l) & 32/128 & \textbf{63.40/62.17} & \textbf{90.28/88.93} & \textbf{38.52/41.49} & \textbf{66.82/68.59} & \textbf{21.11/22.84}\\
     \cmidrule(r){1-8}
    \multicolumn{1}{c|}{\multirow{3}{*}{Cars}} & 
   MoCo v3 (g) & 32/128 & 67.48/68.61 & 90.41/\textbf{91.39}  & 30.16/33.95  & 54.53/58.90  & 9.12/10.01  \\
    & Ours (l) & 32/128 & 64.40/62.44  & 87.91/87.32  & 24.61/31.20  & 47.26/54.20  & 5.47/7.54 \\
    & Ours (g+l) & 32/128 & \textbf{69.73/69.72}  & \textbf{91.92}/91.26 & \textbf{38.37/41.55} & \textbf{64.45/67.24} & \textbf{11.29/13.87}\\
     \cmidrule(r){1-8}
     \multicolumn{1}{c|}{\multirow{3}{*}{CUB}} & 
   MoCo v3 (g) & 32/128 & 71.11/72.75 & 92.32/93.25 & 35.54/42.23 & 64.98/71.14 & 16.18/20.41\\
    & Ours (l) & 32/128 & 72.09/72.28 & 91.53/92.51 & 37.33/43.96 & 63.32/72.02 & 16.94/23.18\\
    & Ours (g+l) & 32/128 & \textbf{72.66/73.23} & \textbf{92.58/93.37} & \textbf{41.82/45.89} & \textbf{67.40/72.75} & \textbf{21.04/25.28}\\  
    \bottomrule
  \end{tabular}
\end{table*}

\subsection{Comparison with Self-supervised Region-level Method}
\label{sec:4.add2}
A comparison of the self-supervised region-level method (EsViT) using a mini-batch size of 128 to evaluate the classification and retrieval performance on the FGVC Aircraft, Stanford Cars, and CUB-200-2011 datasets is shown in Table~\ref{tab:add4}. For FGVC-Aircraft, the Top-1, Top-5, Rank-1, Rank-5 and mAP of the proposed method are 60.52\%, 88.45\%, 36.63\%, 63.37\% and 18.59\%, respectively, which are 5.34\%, 3.09\%, 9.57\%, 10.35\%, and 5.01\% higher than those of EsViT. For Stanford Cars, the Top-1, Top-5, Rank-1 and mAP of the proposed method are 65.47\%, 88.61\%, 32.82\%, and 18.59\%, respectively, which are 6.35\%, 0.77\%, 0.87\%, and 0.03\% higher than EsViT. Although on the CUB-200-2011 dataset, its evaluation metrics are slightly lower than those of EsViT, it has fewer parameters than EsViT.

\subsection{Key Parameter Studies}
\label{sec:4.3}

Masking ratio.
The quantity of valuable fine-grained information is influenced by the masking ratio. A high masking ratio keeps fewer pivotal regions and contains limited learnable valuable information, while a low masking ratio keeps more pivotal regions but introduces less noise information. Image retrieval can evaluate the ability of extractors to capture the most relevant information of images without label intervention. Therefore, we compared only the retrieval performance of our global-local framework when the location-wise mask sampling strategy was used at different masking ratios. Table \ref{tab:3} shows the influence of the masking ratio. For FGVC-Aircraft, the highest Rank-1 is achieved with an 80\% masking ratio, and the best Rank-5 and mAP are observed at a 70\% masking ratio. For Stanford Cars, the highest Rank-1 occurs at a 70\% masking ratio, and the best Rank-5 and mAP are found at a 60\% masking ratio. Overall, a masking ratio of 70\% is a good choice. Moreover, applying a masking ratio of 70\% to FGVC-Aircraft, Stanford Cars and CUB-200-2011 led to an improvement over the baseline model (as shown in Table \ref{tab:2}). To achieve greater benefit, the framework first strikes a balance between noise and valuable information and then maintains as much complementary value as possible. 

\subsection{Ablation Studies}
\label{sec:4.4}
\subsubsection{Impact of branches in the global-local framework}
There are global branches and local branches in our proposed global-local framework. They are evaluated in linear probing and image retrieval with two different mini-batch sizes of 32 and 128, and they employ ViT-S and ViT-B as backbone networks. Both Table \ref{tab:add3} and Table \ref{tab:2} demonstrate that our global-local multi-branch approach is significantly effective.

The second to fourth rows of Table \ref{tab:add3} show the effectiveness of different branches on FGVC-Aircraft. When the mini-batch size is 32, the Top-1, Top-5, Rank-1, Rank-5 and mAP of the global-local method are 54.52\%, 83.29\%, 33.84\%, 60.40\% and 17.26\%, respectively, which are 6.60\%, 3.36\%, 10.72\%, 14.98\%, and 6.80\% higher than those of the global method of MoCo v3. 
The fifth to seventh rows of Table \ref{tab:add3} show the effectiveness of different branches on Stanford Cars. When the mini-batch size is 32, the Top-1, Top-5, Rank-1, Rank-5 and mAP of the global-local method are 58.15\%, 84.44\%, 23.53\%, 44.47\% and 5.88\%, respectively, which are 6.79\%, 4.24\%, 2.04\%, 1.34\%, and 0.38\% higher than those of the global method of MoCo v3. 
The eighth to tenth rows of Table \ref{tab:add3} show the effectiveness of different branches on CUB-200-2011. When the mini-batch size is 32, the Top-1, Top-5, Rank-1, Rank-5 and mAP of the global-local method are 61.82\%, 86.00\%, 36.21\%, 61.93\% and 17.84\%, respectively, which are 1.39\%, 0.74\%, 5.11\%, 3.35\%, and 5.40\% higher than those of the global method of MoCo v3. 

The second to fourth rows of Table \ref{tab:2} show the effectiveness of different branches on FGVC-Aircraft. We can see that not only the global-local method but also the local method led to a significant improvement over MoCo v3. When the mini-batch size is 32, the Top-1, Top-5, Rank-1, Rank-5 and mAP of the global-local method are 63.40\%, 90.28\%, 38.52\%, 66.82\% and 21.11\%, respectively, which are 5.64\%, 3.12\%, 6.15\%, 8.79\%, and 4.39\% higher than those of the global method of MoCo v3. 
The fifth to seventh rows of Table \ref{tab:2} show the effectiveness of different branches on Stanford Cars. Although the performance of the local method is inferior, our global-local method made significant improvements over MoCo v3. When the mini-batch size is 32, the Top-1, Top-5, Rank-1, Rank-5 and mAP of the global-local method are 69.73\%, 91.92\%, 38.37\%, 64.45\% and 11.29\%, respectively, which are 2.25\%, 1.51\%, 8.21\%, 9.92\%, and 2.17\% higher than those of the global method of MoCo v3. 
The eighth to tenth rows of Table \ref{tab:2} show the effectiveness of different branches on CUB-200-2011. Our global-local method also made significant improvements over MoCo v3. When the mini-batch size is 32, the Top-1, Top-5, Rank-1, Rank-5 and mAP of the global-local method are 72.66\%, 92.58\%, 41.82\%, 67.40\% and 21.04\%, respectively, which are 1.55\%, 0.26\%, 6.28\%, 2.42\%, and 4.86\% higher than those of the global method of MoCo v3. 

The improved data on FGVC-Aircraft, Stanford Cars and CUB-200-2011 show that the advantages of the proposed method become more apparent when evaluating retrieval performance; this suggests that the proposed method can focus on more valuable local regions without label intervention. When the mini-batch size increases from 32 to 128, our global-local method still achieves better performance than does MoCo v3.

\begin{figure}
\subfloat[]{\includegraphics[width=0.17\linewidth]{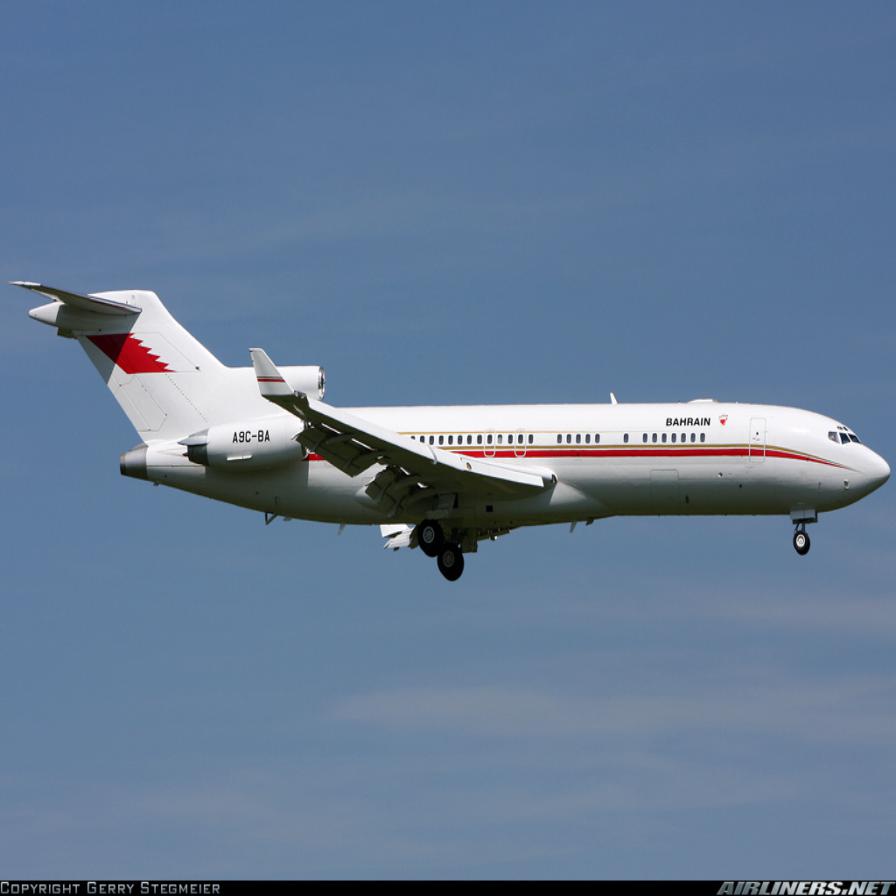}
\label{fig:4-a}}
\hfill
\subfloat[]{ \includegraphics[width=0.17\linewidth]{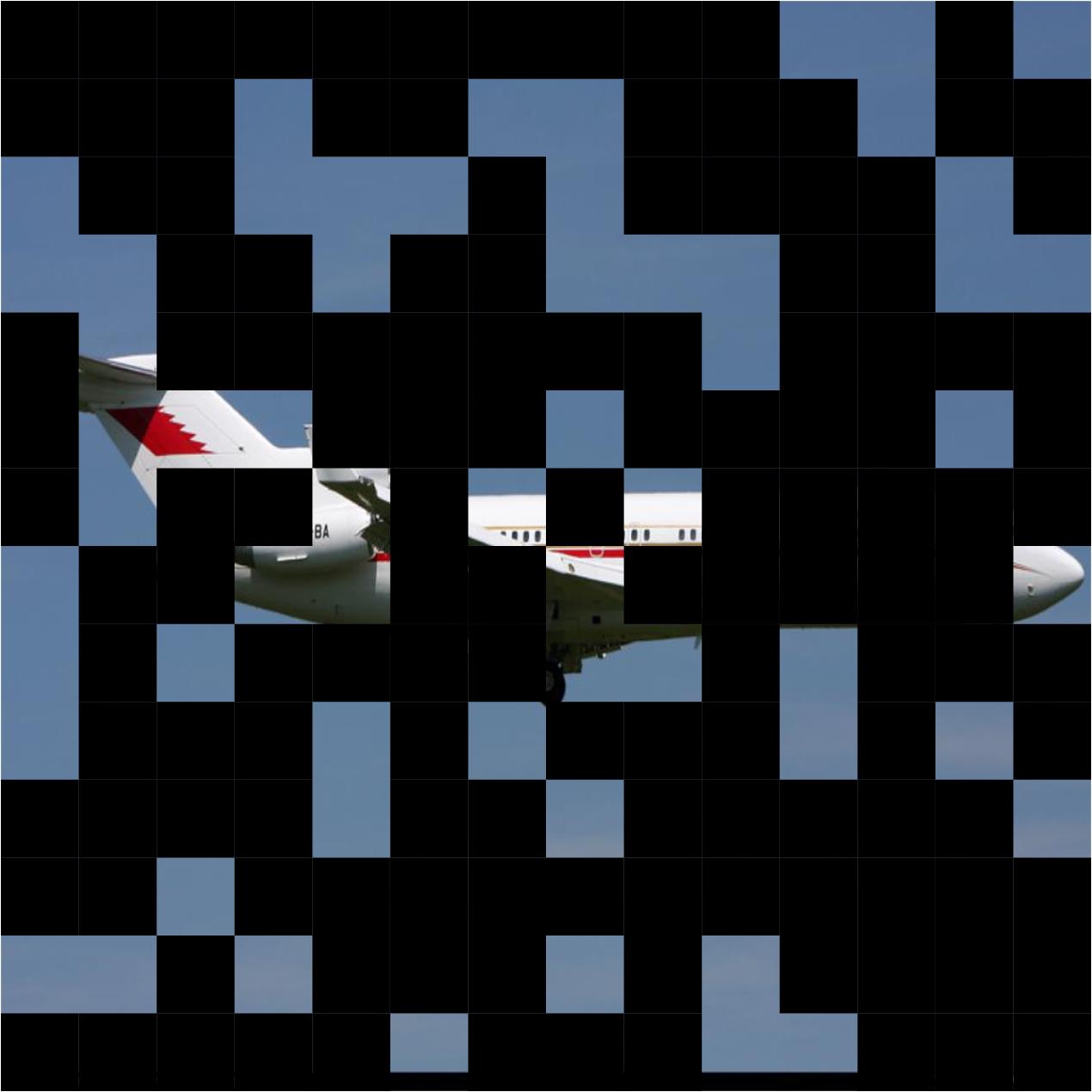}
\label{fig:4-b}}
  \hfill
\subfloat[]{\includegraphics[width=0.17\linewidth]{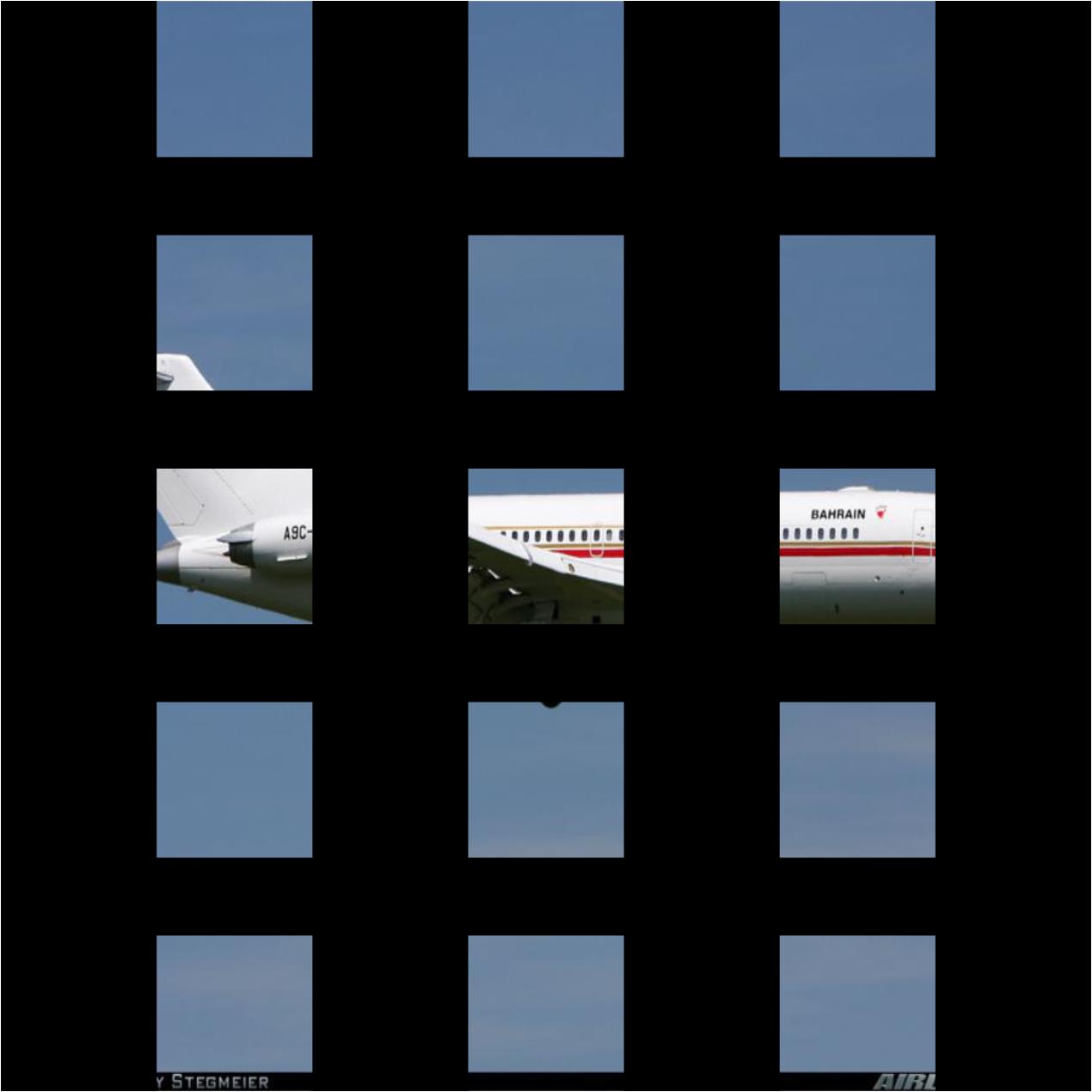}
\label{fig:4-c}}
  \hfill
\subfloat[]{ \includegraphics[width=0.17\linewidth]{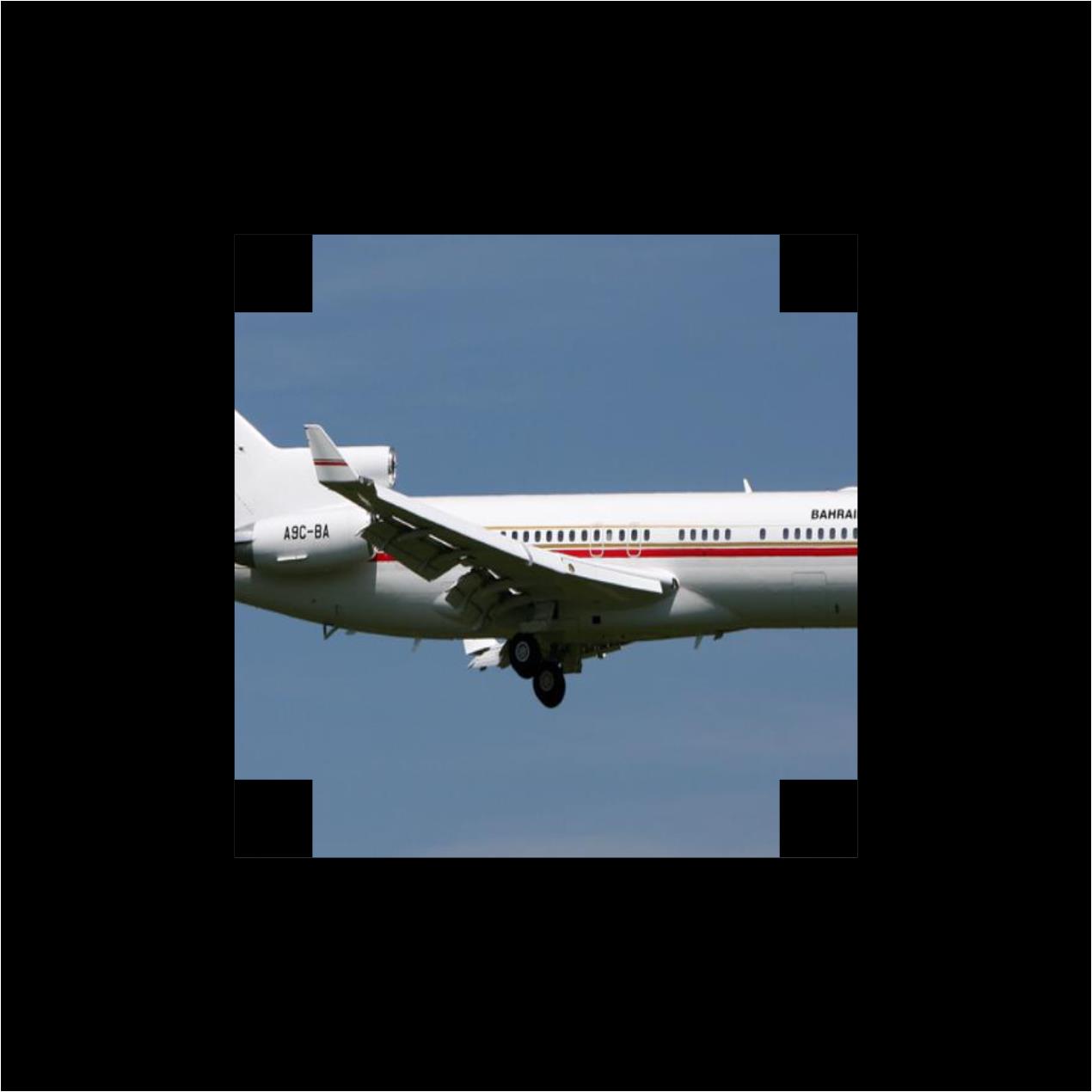}
    \label{fig:4-d}}
  \hfill
  \subfloat[]{ \includegraphics[width=0.17\linewidth]{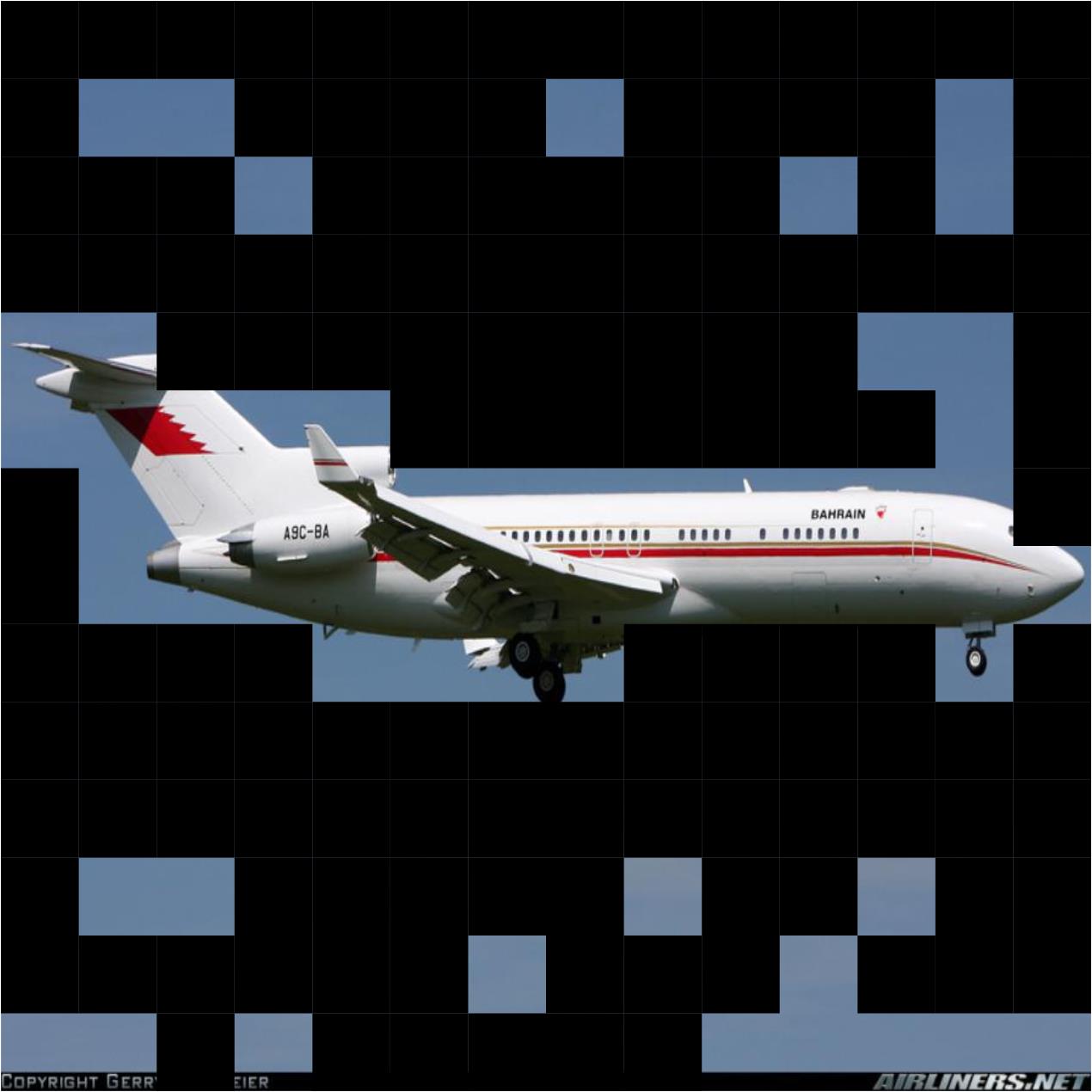}
    \label{fig:4-e}}
  \caption{Mask sampling strategies determine whether local regions are selected, and they influence the quality of the representations extracted by the feature extractor. Each subgraph corresponds to a different strategy. (a) is the original image, (b) shows a random masking strategy, (c) shows a grid-wise masking strategy, (d) shows a border-wise masking strategy, and (e) shows the proposed location-wise masking strategy. Here, the masking ratio for all masking strategies is 70\%.}
  \label{fig:4}
\end{figure}

\begin{figure*}
\subfloat{\includegraphics[width=0.32\linewidth]{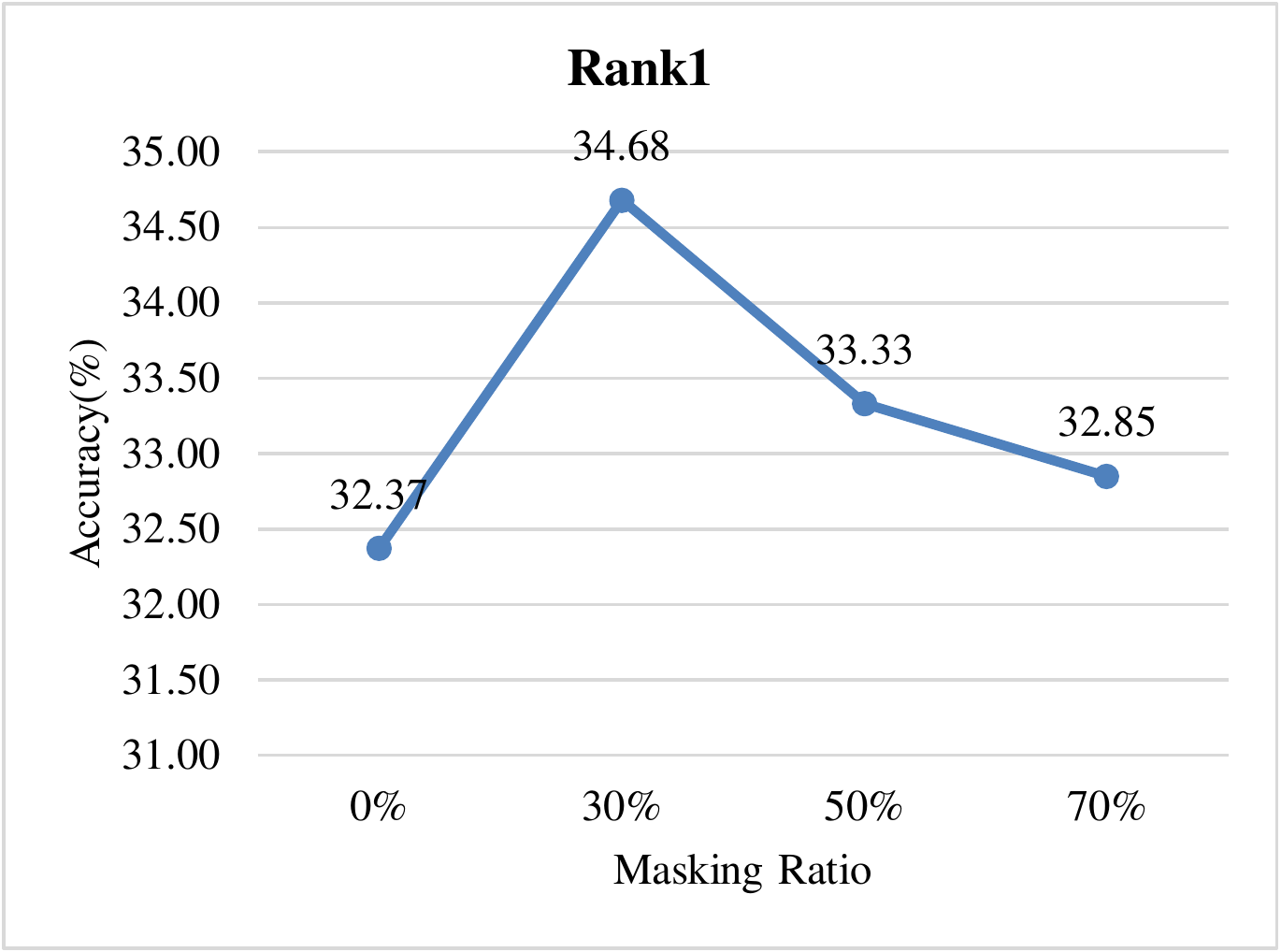}}
\hfill
\subfloat{ \includegraphics[width=0.32\linewidth]{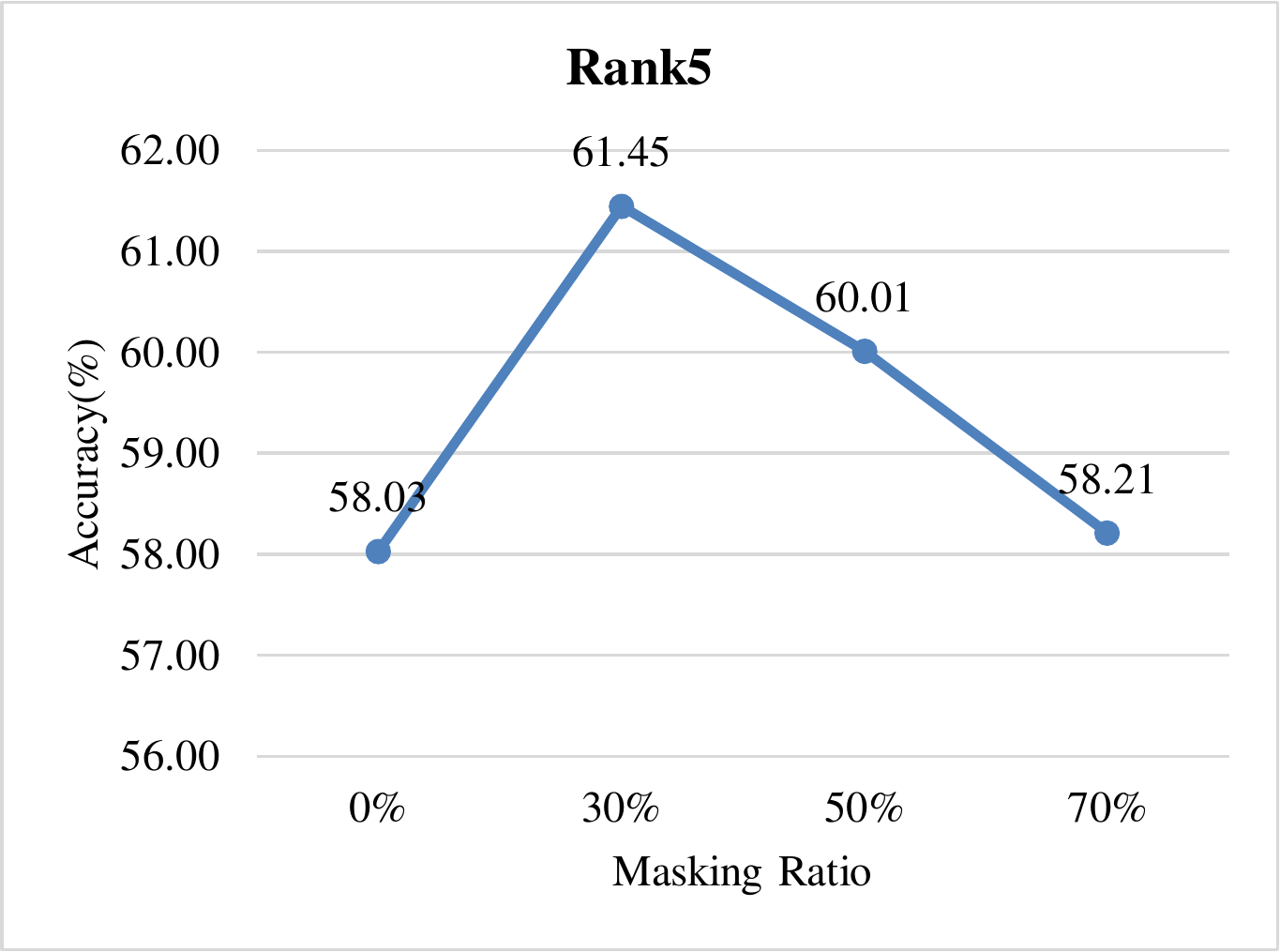}}
  \hfill
\subfloat{\includegraphics[width=0.32\linewidth]{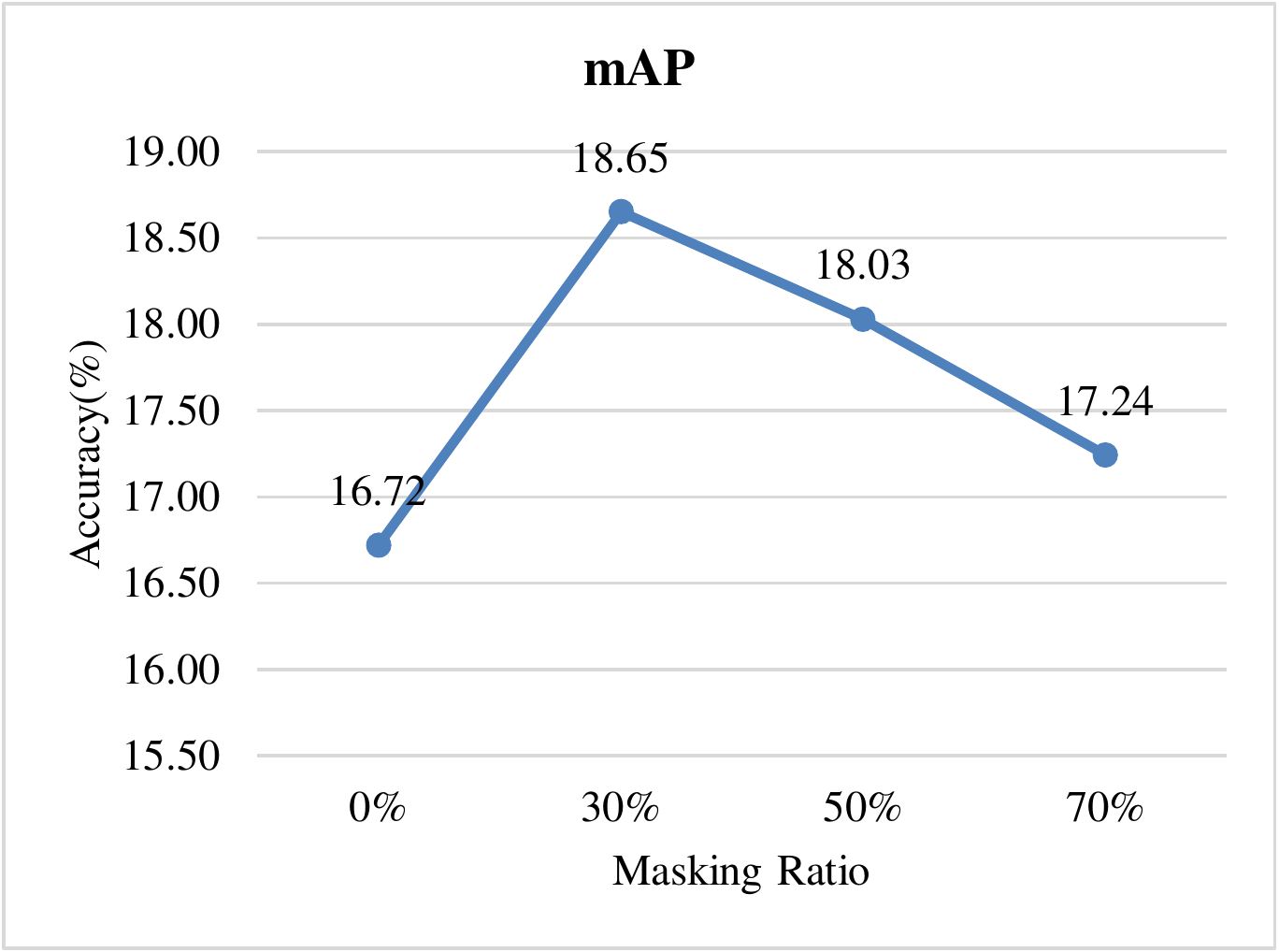}}
  \caption{Performance of the \textbf{border-wise} masking strategy in our global-local framework is evaluated on FGVC-Aircraft at different masking ratios.}
  \label{fig:border}
\end{figure*}

\begin{figure}
  \centering
  \includegraphics[width=0.98\linewidth]{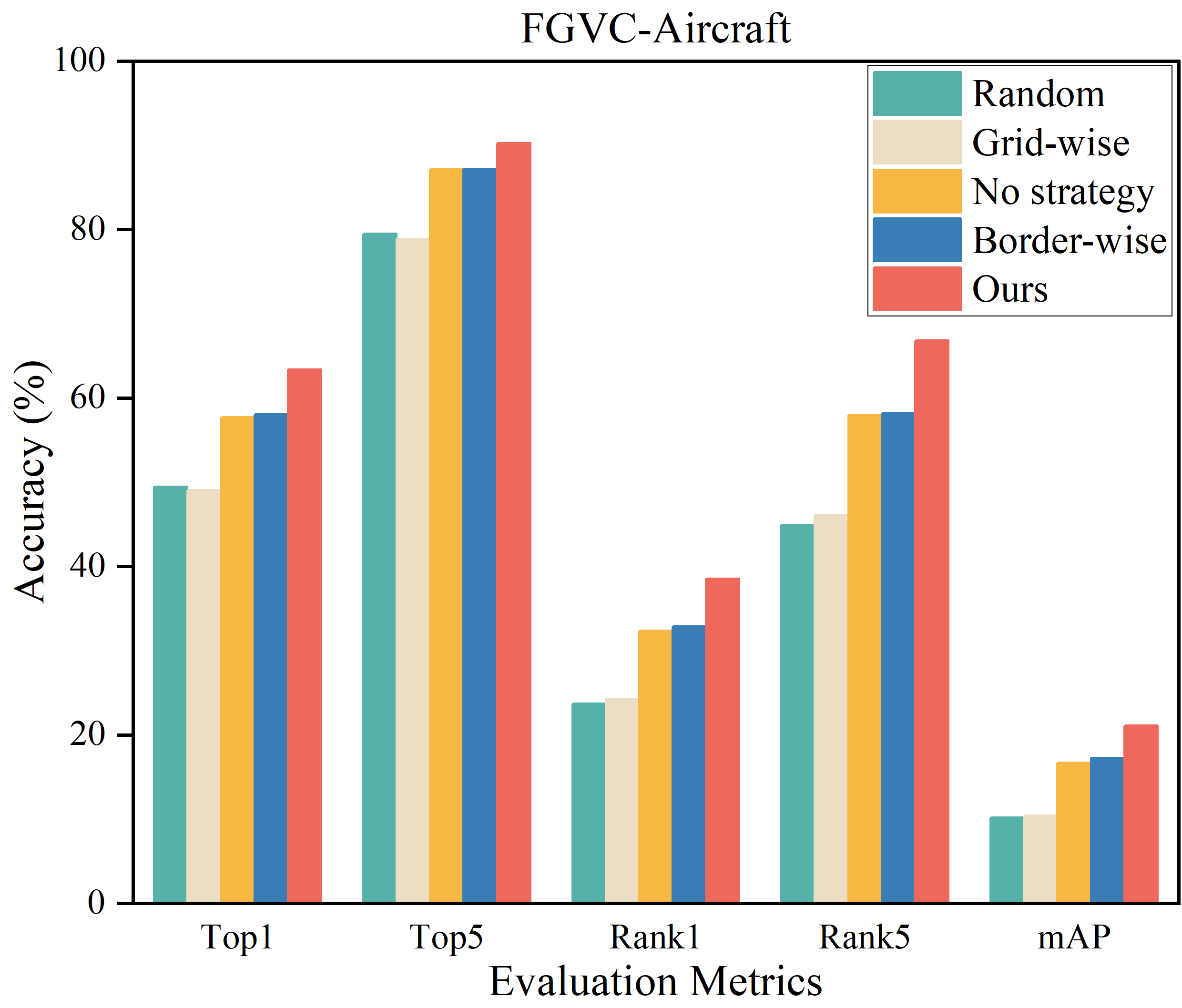}
   \caption{The performance of the global framework without any strategy and our global-local framework with different mask sampling strategies on the FGVC-Aircraft dataset.} 
   \label{fig:5}
\end{figure}

\begin{table}
  \caption{Ablation study on mask sampling strategies for FGVC-Aircraft.}
  \label{tab:4}
  \begin{tabular}{c|cc|ccc}
    \toprule
    \multicolumn{1}{c|}{\multirow{2}{*}{Strategy}} & \multicolumn{2}{c|}{\multirow{1}{*}{Classification}} & \multicolumn{3}{c}{\multirow{1}{*}{Retrieval}} \\
    \cmidrule(r){2-6} & Top-1 & Top-5 & Rank-1 & Rank-5 & mAP \\
    \midrule
     %None & 57.76 & 87.16 & 32.37 & 58.03 & 16.72\\
     Random & 49.48 & 79.51 & 23.70 & 44.91 & 10.20\\
     Grid-wise & 49.09 & 78.88 & 24.27 & 46.14 & 10.44\\
     Border-wise & 58.09 & 87.22 & 32.85 & 58.21 & 17.24\\
     Location-wise (Ours) & \textbf{63.40} & \textbf{90.28} & \textbf{38.52} & \textbf{66.82} & \textbf{21.11}\\
    \bottomrule
  \end{tabular}
\end{table}

\subsubsection{Impact of mask sampling strategy}

In Table \ref{tab:4}, we compare different mask sampling strategies, as illustrated in Fig. \ref{fig:4}. Here, the mini-batch size of all models is set to 32, and we use ViT-B as the backbone network.

The border-wise masking strategy tends to maintain the central region (Fig. \ref{fig:4-d}). Intuitively, this strategy is effective when objects are concentrated mainly at the center of each image in the dataset. Compared with the experimental results without using any strategy (MoCo v3) on the FGVC-Aircraft dataset, the global-local framework with block-wise masking at a ratio of 70\% yields similar experimental results, as shown in Fig. \ref{fig:5}, indicating that the border-wise masking strategy introduces less noise information. As the masking ratio is adjusted, it works reasonably well at a ratio of 30\% and outperforms MoCo v3, as shown in Fig. \ref{fig:border}. In this sense, adding local branches is favorable.

The random masking strategy removes a certain number of patches without rules (Fig. \ref{fig:4-b}). The grid-wise \cite{mae} masking strategy regularly keeps one of every four patches (Fig. \ref{fig:4-c}). Both the random and grid-wise masking strategies achieve worse performance and cannot effectively improve the accuracy of the base model. These two strategies are ineffective because they ignore the semantic information of the image and introduce a large amount of noise information.

The proposed location-wise masking strategy (Fig. \ref{fig:4-e}) based on attention weights works the best for our global-local framework. It keeps the patches with higher attention weights, which helps the model learn valuable representations. Both linear probing and image retrieval achieve meaningful improvements. Accordingly, the proposed strategy preserves the global semantic information of the image while also extracting valuable insights.

\begin{figure*}
\subfloat{\includegraphics[width=0.32\linewidth]{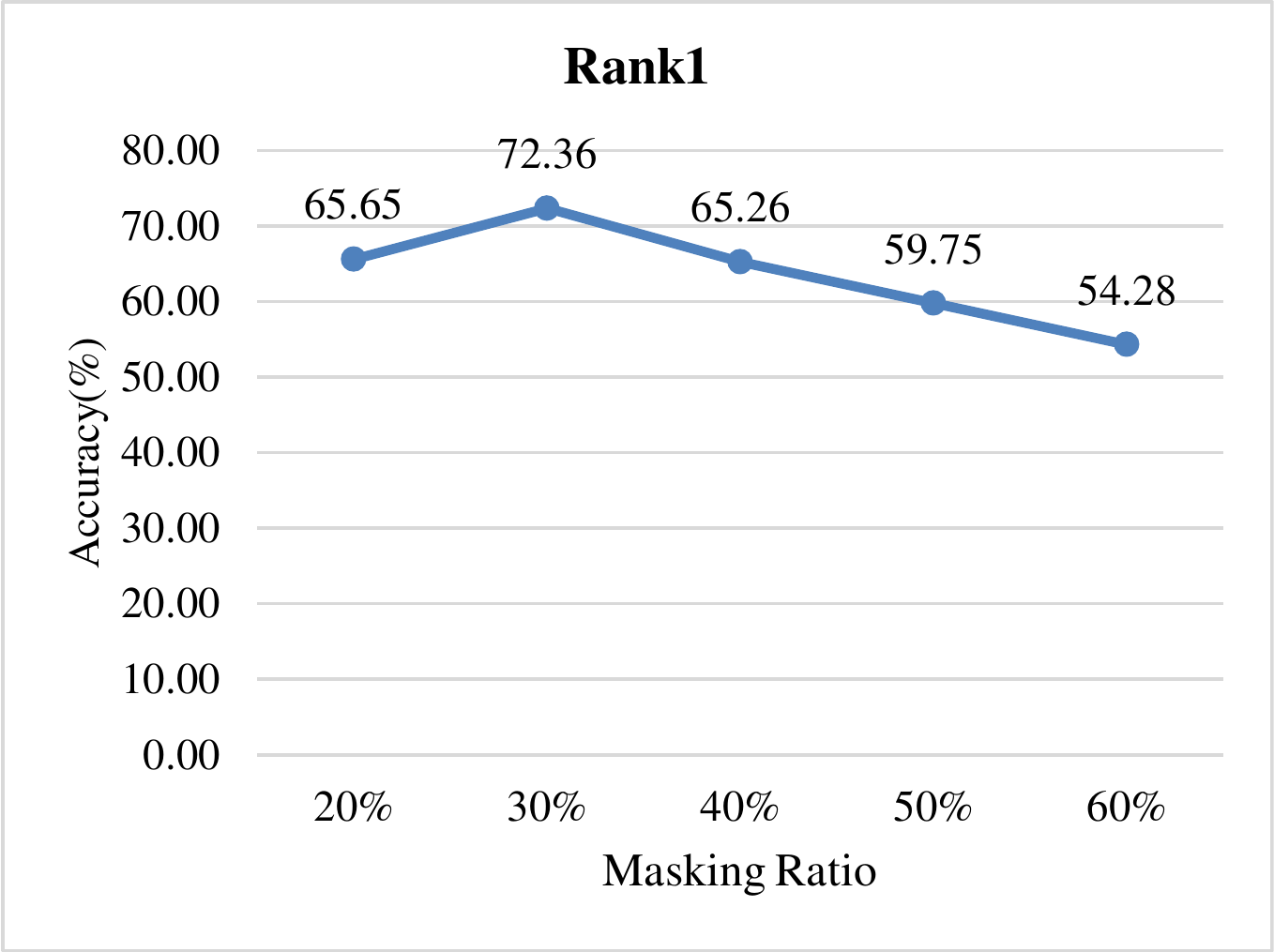}}
\hfill
\subfloat{ \includegraphics[width=0.32\linewidth]{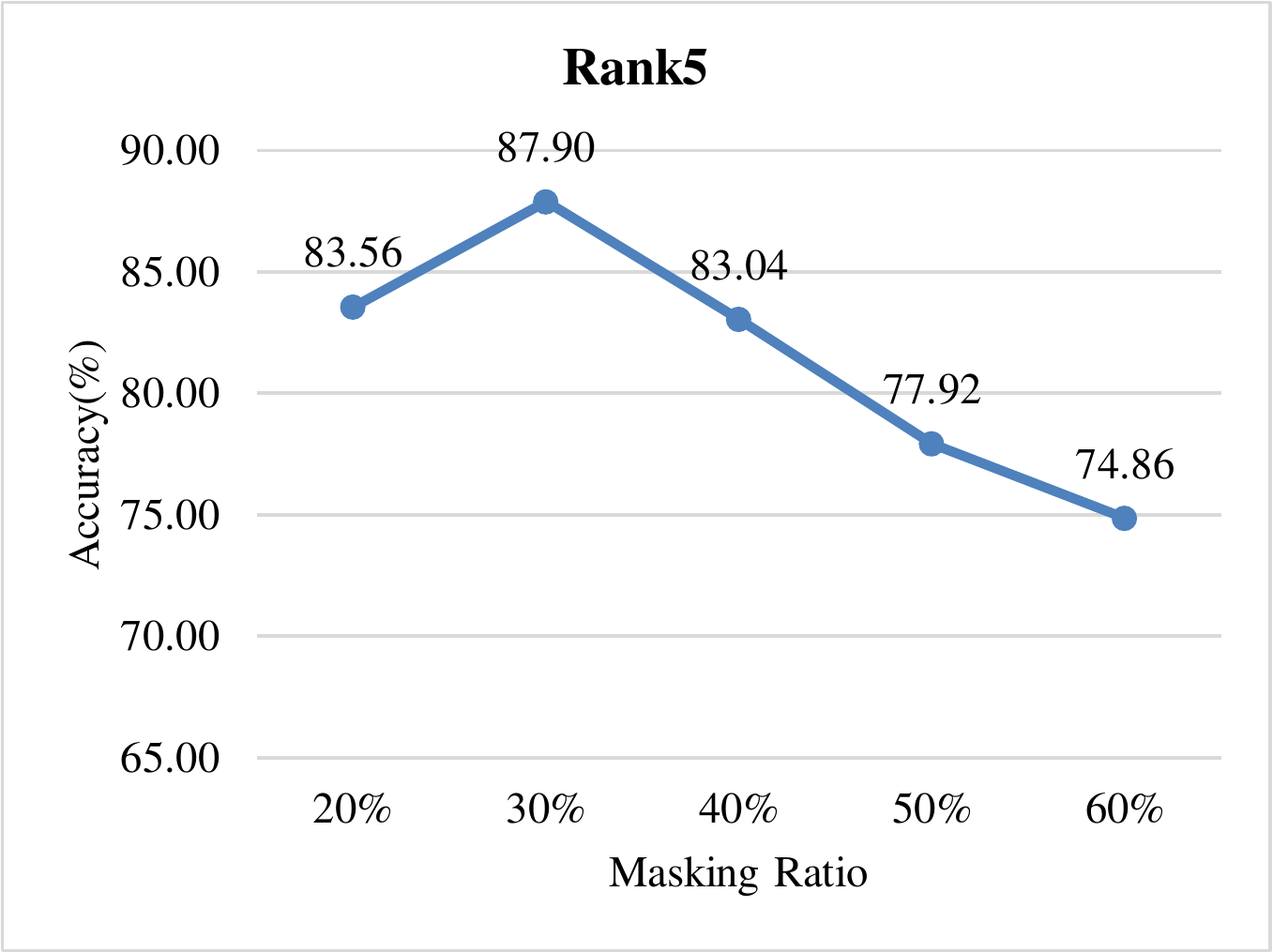}}
  \hfill
\subfloat{\includegraphics[width=0.32\linewidth]{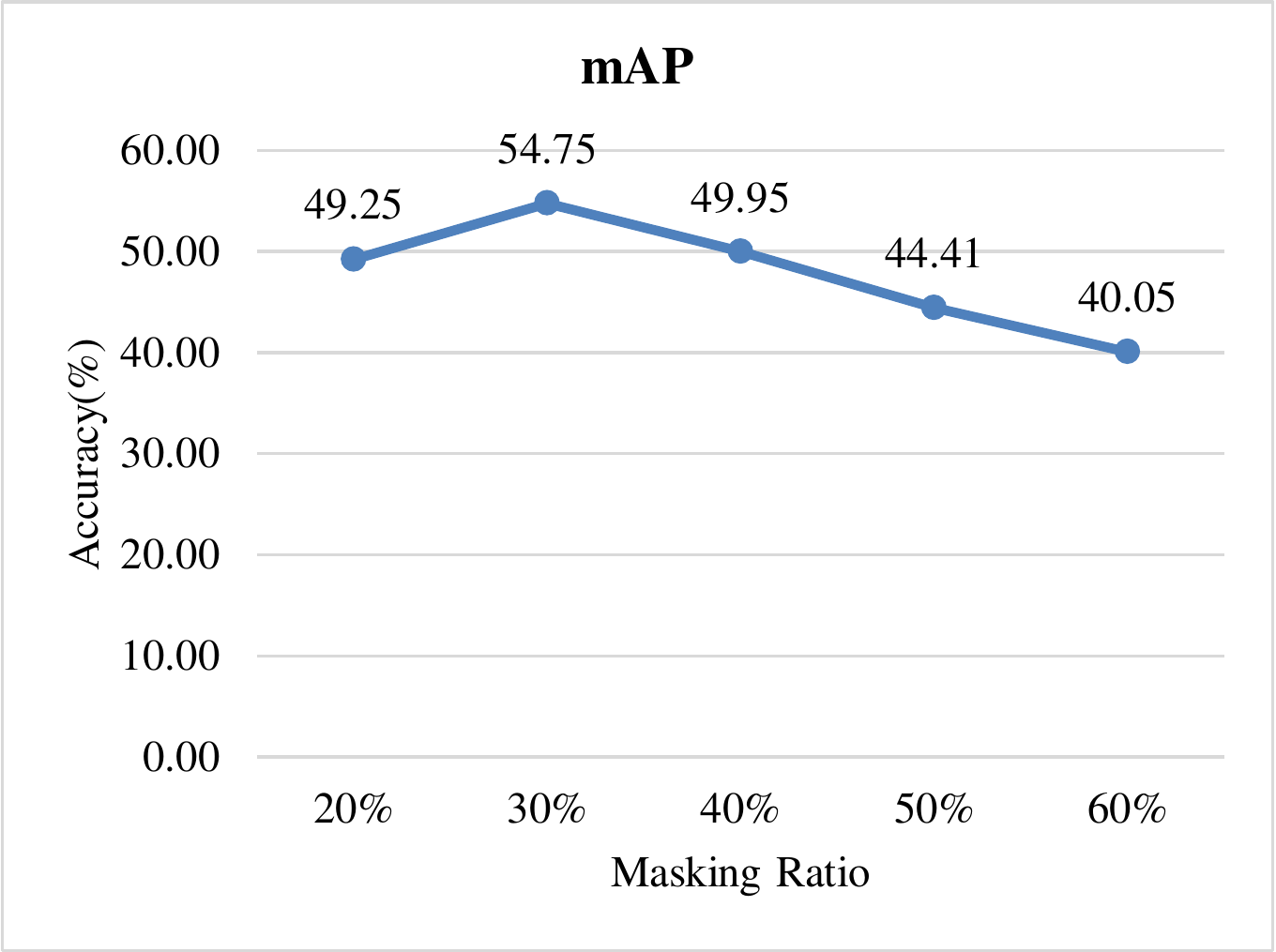}}
  \caption{Performance of different masking ratios on Caltech-101. The masking ratio (30\%) works well.}
  \label{fig:6}
\end{figure*}

\begin{table}
  \caption{Classification and retrieval performance of our method evaluated on Caltech-101. }
  \label{tab:6}
  \centering
  \begin{tabular}{c|cc|ccc}
    \toprule
    \multicolumn{1}{c|}{\multirow{2}{*}{Method}} & \multicolumn{2}{c|}{\multirow{1}{*}{Classification}} & \multicolumn{3}{c}{\multirow{1}{*}{Retrieval}} \\
    \cmidrule(r){2-6} & Top-1 & Top-5 & Rank-1 & Rank-5 & mAP \\
    \midrule
     MoCo v3 (g) & 58.63 & 81.83 & 58.16 & 76.75 & 42.79\\
     Ours (l) & 48.60 & 73.18 & 50.28 & 69.01 & 35.04\\
     Ours (g+l) & \textbf{79.38} & \textbf{95.27} & \textbf{72.36} & \textbf{87.90} & \textbf{54.75}\\
    \bottomrule
  \end{tabular}
\end{table}

\begin{figure}
\centering
% \begin{picture}(0,0)
%     \put(25,75){Original}
%     \put(98,75){MoCo v3}
%     \put(180,75){Ours}
%     \put(265,75){Original}
%     \put(338,75){MoCo v3}
%     \put(420,75){Ours}
%  \end{picture}
\subfloat[FGVC-Aircraft]{\includegraphics[width=0.48\linewidth]{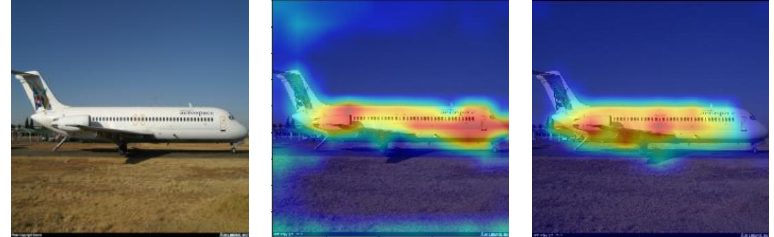}}
\hfill
\subfloat[Stanford Cars]{ \includegraphics[width=0.48\linewidth]{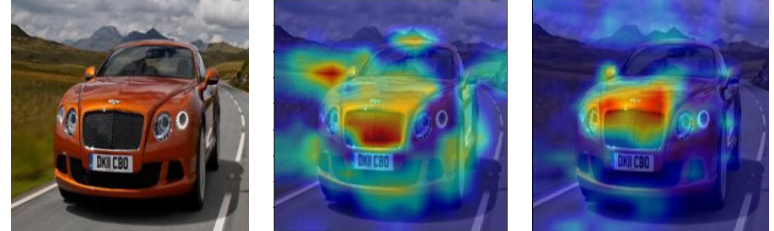}}
  \hfill
\subfloat[CUB-200-2011]{\includegraphics[width=0.48\linewidth]{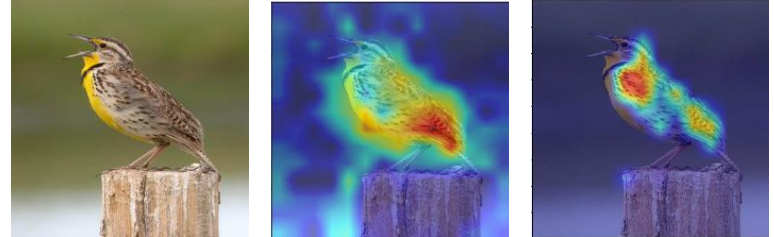}}
  \hfill
\subfloat[Caltech-101]{\includegraphics[width=0.48\linewidth]{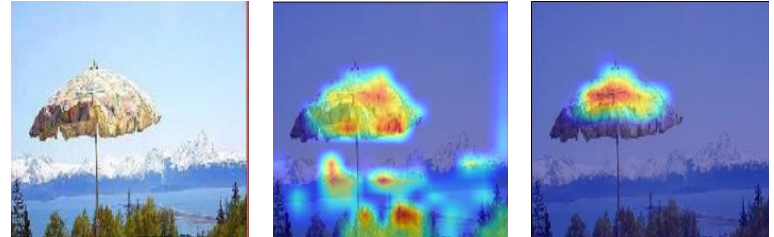}}
  \caption{Grad-CAM visualization for MoCo v3 and our method on the FGVC-Aircraft, Stanford
 Cars, CUB-200-2011 and Caltech-101 datasets. In each subgraph, the first column displays the original image, the second column displays the Grad-CAM results for MoCo v3, and the third column displays the Grad-CAM results for our global-local method. Compared with the global method MoCo v3, which focuses on all regions of the image, including the background, our method excels at discerning features from the object. }
\label{fig:compare}
\end{figure}

\begin{figure}
  \centering
%  \begin{picture}(0,0)
%    \put(15,-15){Original}
%    \put(67,-15){Pivotal regions}
%    \put(142,-15){Grad-CAM}
%    \put(214,-15){Original}
%    \put(266,-15){Pivotal regions}
%    \put(341,-15){Grad-CAM}
% \end{picture}
  \includegraphics[width=0.98\linewidth]{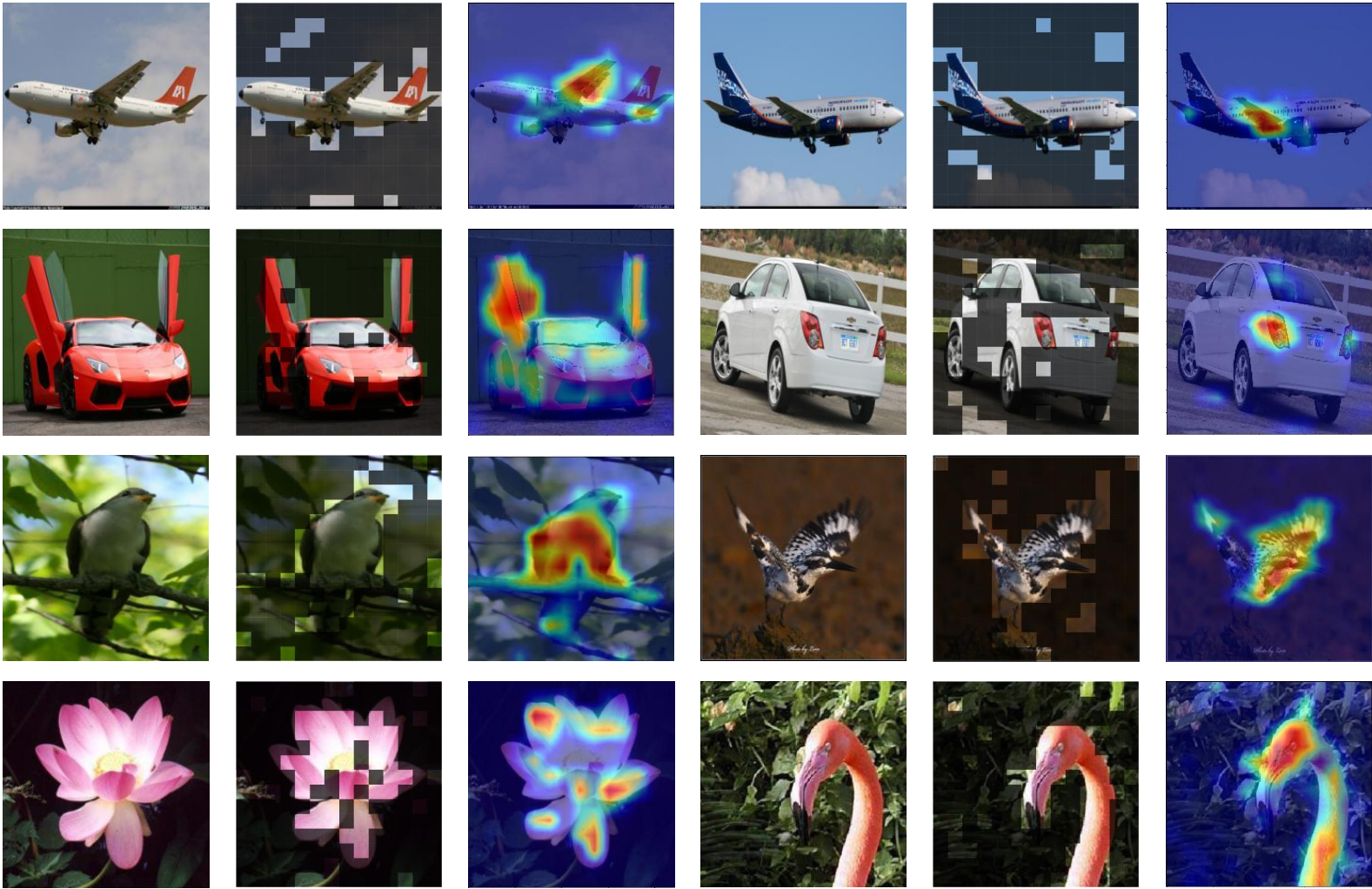}
   \caption{The original images (first and fourth columns), pivotal regions (second and fifth columns) and Grad-CAM (third and sixth columns) visualization for our method on the FGVC-Aircraft (first row), Stanford Cars (second row), CUB-200-2011 (third row), and Caltech-101 (fourth row) datasets. The noncritical regions of the image are locally masked, leaving the pivotal regions exposed. The learning features are enhanced in pivotal regions of the object.}
   \label{fig:7}
\end{figure}

\subsection{Performance on General Object Recognition}
\label{sec:4.5}
Considering that important local information is helpful for general object recognition tasks, we also investigate the proposed global-local framework on Caltech-101, which is a general visual dataset. Similarly, it is evaluated in linear probing and image retrieval with a mini-batch size of 64. We researched for a suitable masking ratio on the generic visual dataset Caltech-101. As shown in Fig. \ref{fig:6}, the best masking ratio is 30\%. This is due to the differences in the characteristics between fine-grained visual datasets and general image datasets. Compared with fine-grained visual datasets, general visual datasets consist of more diverse and distinct object categories, where a lower masking ratio is sufficient to filter out the nontarget information and capture the overall structure of the images. The experimental results are described in Table \ref{tab:6}. Although the performance of the local method is not as good as that of the global method MoCo v3, the best performance is achieved when the two cooperate with each other, which shows that the local method's pretext task LoDisc is beneficial for discovering supplementary information and leads to better cooperative results in terms of object recognition. Additionally, the Top-1, Top-5, Rank-1, Rank-5 and mAP of the global-local method are 79.38\%, 95.27\%, 72.36\%, 87.90\% and 54.57\%, respectively, which are 20.75\%, 13.44\%, 14.20\%, 11.15\%, and 11.96\% higher than those of the global method MoCo v3. This finding indicates that our global-local self-supervised contrastive framework is also effective in general object recognition.

\subsection{Visualization of Attention Maps}
\label{sec:4.6}

The attention maps are visualized in Fig. \ref{fig:compare} and Fig. \ref{fig:7} via Grad-CAM \cite{Gradcam}. Fig. \ref{fig:compare} shows that, compared with the baseline method, our method directs the model's attention more toward the foreground of the object and even toward local parts of the object; this leads to reduced susceptibility to interference from the image background, as shown in the third and sixth columns of the figure. 
Fig. \ref{fig:7} clearly shows that the model's attention is guided by pivotal regions, which facilitates the discovery of discriminative features related to the semantics of the image within these regions. Overall, our global-local method successfully captures the more discriminative regions for an object, i.e., lights or doors for cars, as shown in the second line of Fig. \ref{fig:7}, and belly or wings for birds, as shown in the third line of Fig. \ref{fig:7}; this is the secret to significantly enhancing model performance and further demonstrates the effectiveness and scientificity of our method.

\section{Conclusions}
In this paper, a pure self-supervised global-local fine-grained contrastive learning framework is proposed to learn discriminative features at both the global and crucial local levels. To obtain subtle local features, a novel pretext task called local discrimination is developed to explicitly supervise the self-supervised model's focus toward important local regions, which depends on a simple but effective location-wise mask sampling strategy. The proposed method achieves state-of-the-art performance in classification and retrieval tasks for fine-grained visual recognition problems; it is also effective in general object recognition. We show that the local discrimination pretext task can offer valuable insights for identifying discriminative local regions without changing semantic information, and the global-local framework further learns the refined feature representations of images that are beneficial for fine-grained visual recognition. 

Self-supervised fine-grained recognition offers a promising direction for alleviating the reliance on manual annotations. However, the supervisory signals derived from the pretext tasks in self-supervised models still fall short of the precision provided by manually annotated labels, contributing to the performance gap between supervised and self-supervised fine-grained recognition. Narrowing this gap remains a key challenge for future research in self-supervised fine-grained representation learning. Our work highlights the potential of applying self-supervised learning to fine-grained visual recognition tasks, and we plan to further improve its performance and expand its applicability to a broader range of fine-grained tasks through continued research.

% {\appendix[Proof of the Zonklar Equations]
% Use $\backslash${\tt{appendix}} if you have a single appendix:
% Do not use $\backslash${\tt{section}} anymore after $\backslash${\tt{appendix}}, only $\backslash${\tt{section*}}.
% If you have multiple appendixes use $\backslash${\tt{appendices}} then use $\backslash${\tt{section}} to start each appendix.
% You must declare a $\backslash${\tt{section}} before using any $\backslash${\tt{subsection}} or using $\backslash${\tt{label}} ($\backslash${\tt{appendices}} by itself
%  starts a section numbered zero.)}


\newpage

\section*{Biography Section}
\vspace{-33pt}
{\begin{IEEEbiography}
[{\includegraphics[width=1in,height=1.25in,clip,keepaspectratio]{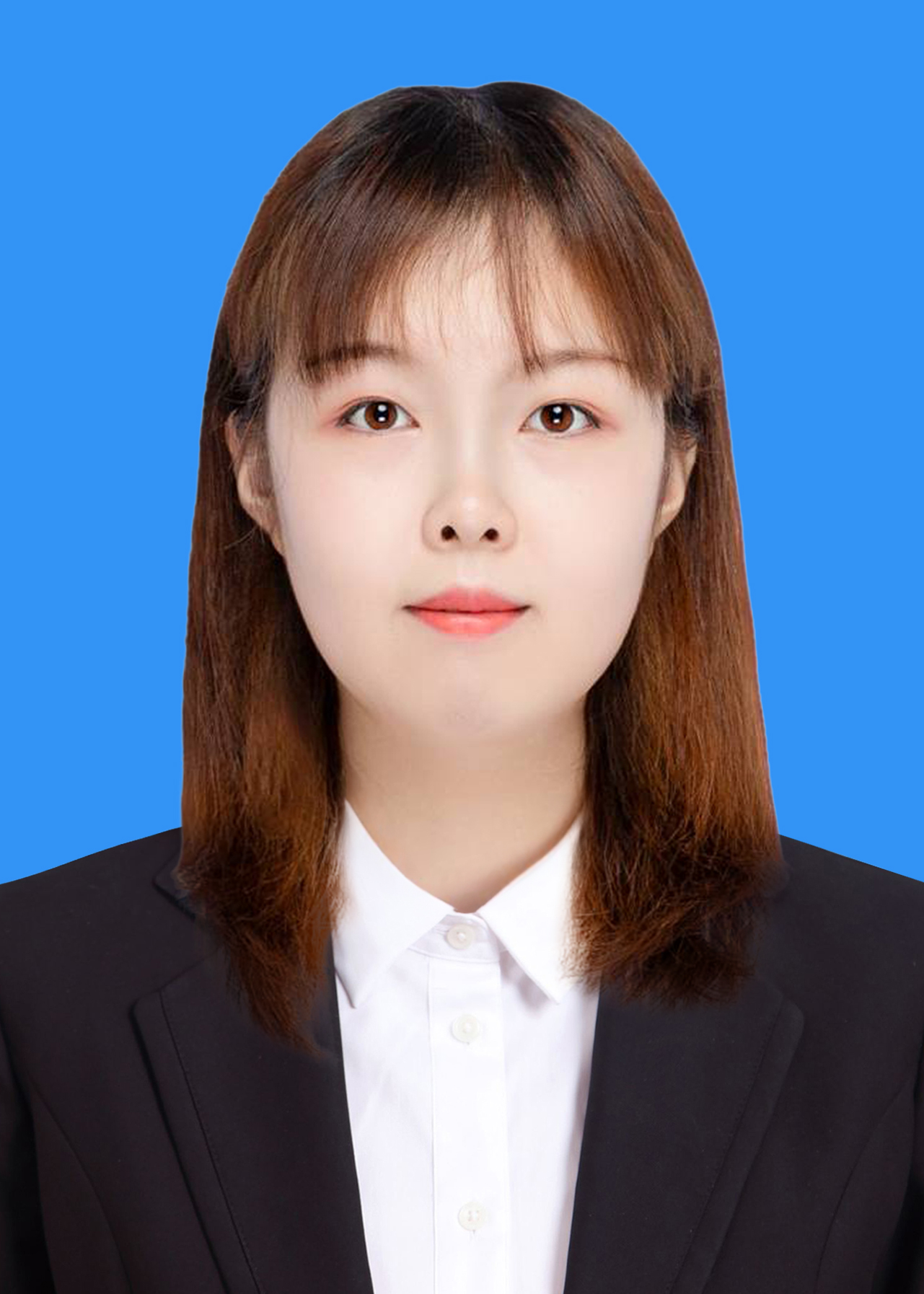}}]{Jialu Shi} is currently a master student of the Ocean University of China, Qingdao, China. She graduated from Yunnan Minzu University in 2022 with an undergraduate degree. Her research interests include computer vision and self-supervised learning.
\end{IEEEbiography}}
\vspace{-33pt}
{\begin{IEEEbiography}
[{\includegraphics[width=1in,height=1.25in,clip,keepaspectratio]{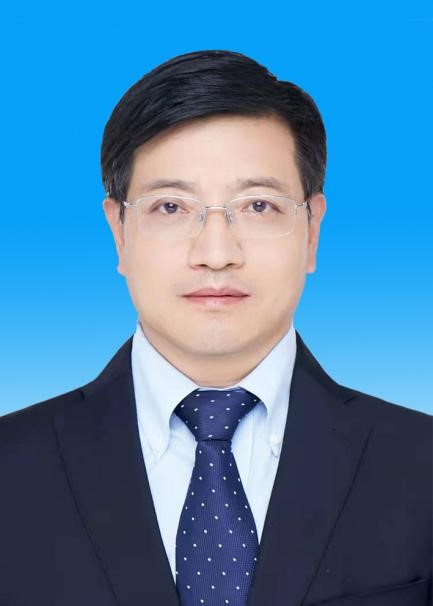}}]{Zhiqiang Wei} is currently a professor with the Ocean University of China, Qingdao, China. He received the Ph.D. degree from Tsinghua University, Beijing, China, in 2001. His current research interests are in the fields of intelligent information processing, computer vision and big data analytics.
\end{IEEEbiography}}
\vspace{-33pt}
{\begin{IEEEbiography}
[{\includegraphics[width=1in,height=1.25in,clip,keepaspectratio]{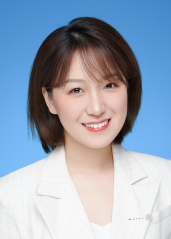}}]{Jie Nie} is currently a professor in Ocean University of China where she received the B.S. and Ph.D. degrees from in 2002 and 2011, respectively, all in computer science. She was a visiting scholar in University of Pittsburgh, USA from Sept. 2009 to Sept. 2010. After that she was a postdoctoral fellow with Tsinghua University from 2015 to 2017, Beijing, China. Her current research interests are in the fields of artificial intelligence and visual analysis of marine big data.
\end{IEEEbiography}}
\vspace{-33pt}
{\begin{IEEEbiography}
[{\includegraphics[width=1in,height=1.25in,clip,keepaspectratio]{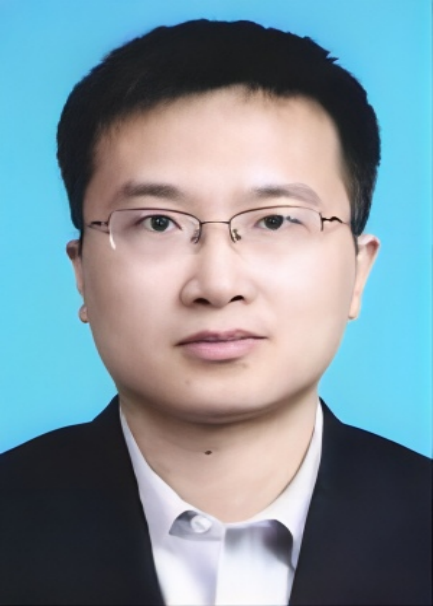}}]{Lei Huang}  is currently a professor with the Ocean University of China, Qingdao, China. He received
 the Ph.D. degree in computer science from the Institute of Computing Technology, Chinese Academy of Sciences, Beijing, China, in 2013. His current research interests are in the fields of multimedia content analysis and retrieval, computer vision, machine learning and marine big data analysis.
\end{IEEEbiography}}

\vfill

\end{document}